%% file: main.tex
\documentclass[10pt,twocolumn,letterpaper]{article}

\usepackage{cvpr}              

\input{preamble}

\definecolor{cvprblue}{rgb}{0.21,0.49,0.74}
\usepackage[pagebackref,breaklinks,colorlinks,allcolors=cvprblue]{hyperref}

\def\confName{CVPR}
\def\confYear{2027}

\title{QuantileGS: Quantile-Based Conformal Bounds on RGB Rendering Errors for Novel View Synthesis with 3D Gaussian Splatting}

\author{%
Haiyang Xu\textsuperscript{1,\textdagger}\quad Ziye Zhang\textsuperscript{1,\textdagger}\quad Bin Pan\textsuperscript{1,*}\quad Xia Xu\textsuperscript{2}\quad Zhenwei Shi\textsuperscript{3}\\[6pt]
{\small\textsuperscript{1}School of Statistics and Data Science, LEBPS, KLMDASR, LPMC, AAIS, NITFID, Nankai University}\\[2pt]
{\small\textsuperscript{2}School of Computer Science and Technology, Tiangong University}\\[2pt]
{\small\textsuperscript{3}Department of Aerospace Intelligent Science and Technology, School of Astronautics, Beihang University}\\[6pt]
{Project Page: \url{https://quantilegs.github.io}}
}

\begin{document}
\maketitle
\cvprauthornote

\input{sec/0_abstract}    
\input{sec/1_intro}
\input{sec/2_related_work}
\input{sec/3_preliminaries}
\input{sec/4_quantile}
\input{sec/5_experiments}
\input{sec/6_conclusion}
\input{sec/llm_usage_statement}
\input{sec/acknowledgment}


{
    \small
    \bibliographystyle{ieeenat_fullname}
    \bibliography{main}
}

\input{sec/X_suppl}

\end{document}

%% file: preamble.tex
\usepackage{graphicx}
\usepackage{xcolor}
\usepackage{colortbl}
\usepackage{booktabs,multirow,array,tabularx,siunitx}
\usepackage{amsmath,amsthm,bm}
\usepackage[normalem]{ulem}
\usepackage{float}
\usepackage{algorithm,algpseudocode}
\usepackage{cuted}
\usepackage{dsfont}
\newtheorem{qgstheorem}{Theorem}
\definecolor{bestcolor}{HTML}{BDAACB}
\definecolor{secondcolor}{HTML}{D6CADF}
\definecolor{thirdcolor}{HTML}{EAE4EF}

\providecommand{\best}[1]{\cellcolor{bestcolor}\textbf{#1}}
\newcommand{\qgsSecond}[1]{\cellcolor{secondcolor}#1}
\providecommand{\third}[1]{\cellcolor{thirdcolor}#1}

\colorlet{qgsSummary}{thirdcolor!65!white}

\makeatletter

\renewcommand{\footnoterule}{%
    \kern -3pt
    \hrule \@width 0.45\columnwidth
           \@height 0.4pt
    \kern 3pt
}

\newcommand{\cvprauthornote}{%
    \begingroup
    \renewcommand{\thefootnote}{}%
    \renewcommand{\@makefntext}[1]{%
        \noindent ##1%
    }%
    \footnotetext{%
        \setlength{\parindent}{0pt}%
        \raggedright
        \hspace*{2em}\textsuperscript{\textdagger}\,Equal contribution. \par
        \hspace*{2em}\textsuperscript{*}\,Corresponding author.\par
       
    }%
    \endgroup
}

\makeatother

%% file: sec/0_abstract.tex
\begin{abstract}
Uncertainty quantification for novel view synthesis with 3D Gaussian Splatting (3DGS) assesses rendering reliability at unseen viewpoints. This reliability depends on the discrepancy between rendered and true colors, which cannot be measured without ground-truth images. We therefore quantify uncertainty in this unknown discrepancy through upper bounds on per-pixel RGB mean absolute error. However, learned bounds may underestimate these errors and overstate rendering reliability, motivating calibration to a prescribed coverage level. In this paper, we propose QuantileGS, a post-hoc framework that combines learned rendering-error quantiles with scale-normalized conformal calibration to produce spatially adaptive error bounds. The framework comprises three phases.
Phase~1, Rendering-Error Computation, measures RGB mean absolute error between observed images and predictions from a frozen 3DGS model. Phase~2, Ordered Gaussian Quantile Fields, learns two scalar attributes per Gaussian to predict lower and upper
rendering-error quantiles. Phase~3, Scale-Normalized Conformal Calibration, calibrates
the upper prediction on held-out images, using the quantile gap to scale the correction.
Under the stated assumptions, the resulting bounds provide finite-sample marginal coverage.
At a nominal coverage level of $90\%$, calibration raises mean empirical test coverage from $69.6\%$ to $89.6\%$ on LLFF and from $80.1\%$ to $92.1\%$ on Mip-NeRF~360.
On LF, coverage remains conservative at $96.3\%$ before and $96.4\%$ after calibration.
Applications to active view selection, Gaussian pruning, and RGB-D simultaneous localization and mapping illustrate how these bounds inform downstream decisions.
\end{abstract}

%% file: sec/1_intro.tex
\section{Introduction}
\label{sec:intro}

\begin{figure*}[t]
    \centering
    \includegraphics[width=\textwidth]{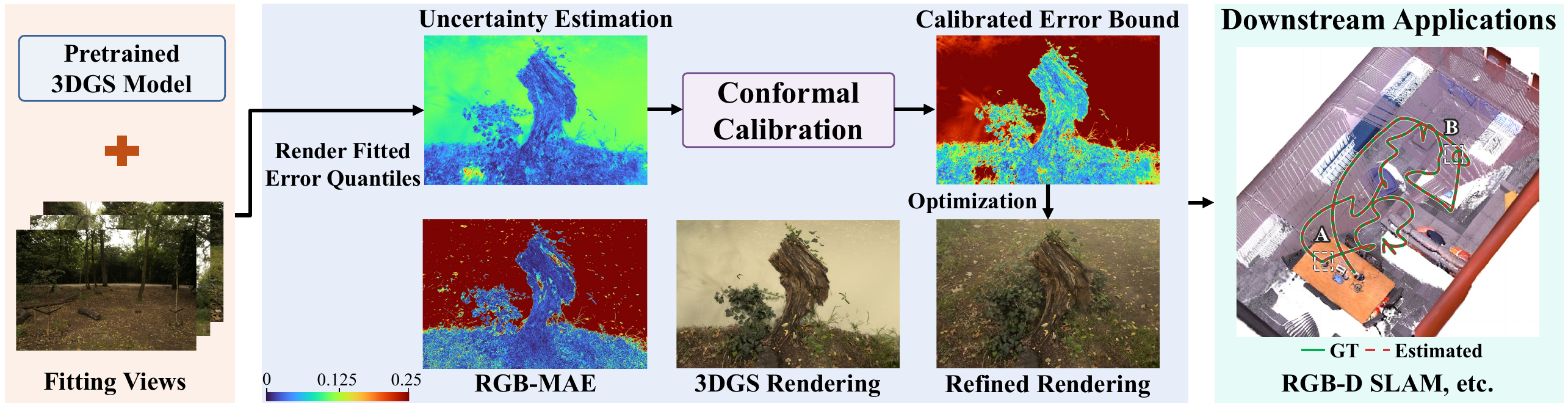}
    \caption{QuantileGS estimates rendering-error uncertainty and uses conformal calibration to produce pixel-wise RGB-MAE upper bounds. The reference error map compares the original 3DGS rendering with ground-truth RGB. Calibrated UQ guides model optimization to produce the refined rendering. We apply the resulting bounds to RGB-D SLAM, as well as active view selection and Gaussian pruning.
    }
    \label{fig:quantilegs_teaser}
\end{figure*}

Uncertainty quantification (UQ) for novel view synthesis (NVS) with 3D Gaussian Splatting (3DGS)~\cite{kerbl20233d,bao20253d} assesses the reliability of images rendered from previously unseen viewpoints. Despite advances in generative view synthesis~\cite{chan2023generative}, efficient rendering~\cite{yariv2023bakedsdf}, and sparse-view
reconstruction~\cite{zhu2024fsgs,zheng2025nexusgs,zhao2025selfensembling},
high reconstruction quality on observed views does not ensure accurate predictions at unseen viewpoints. Occlusion, limited observations, and view-dependent appearance leave scene properties weakly constrained, producing errors that vary across pixels and viewpoints. The reliability of a rendered pixel depends on its agreement with the ground-truth RGB at the requested viewpoint, which is unavailable at inference. We therefore characterize the predictive uncertainty of a fixed 3DGS model through the RGB errors its predictions may incur. Specifically, we estimate upper quantiles of per-pixel RGB mean absolute error (RGB-MAE), the average absolute discrepancy across the three color channels. This expresses uncertainty in terms of how large the unknown rendering error may be.

Existing methods estimate rendering uncertainty through parameter
variation~\cite{li2024variational,lyu2024manifold,wu2026horseshoe,jia2026renderingaware}, rendering sensitivity~\cite{jiang2024fisherrf}, and cross-view
consistency~\cite{safadoust2026warprf}. Others use reconstruction residuals and observation coverage to predict errors or identify unreliable
regions~\cite{galappaththige2026predictive,gottwald2026primu,ling2025ocsplats}.
These approaches characterize different aspects of rendering reliability.
Prediction variability and rendering sensitivity describe how outputs change, but do not by themselves measure agreement with ground truth. For example, predictions from different model samples can agree closely while sharing a systematic color bias. Learning from reconstruction residuals connects uncertainty more directly to observed RGB discrepancies, providing a basis for interpreting the reliability of a fixed renderer in terms of its prediction errors.

However, these methods do not generally provide RGB-MAE upper bounds at a prescribed reliability level. Learned upper error quantiles characterize large deviations
that mean-error estimates may miss, but need not attain their nominal levels on unseen observations, motivating calibration with held-out errors~\cite{angelopoulos2023conformal,angelopoulos2025learn}.
Coverage is the probability that a new pixel's RGB-MAE does not exceed its predicted bound, with tightness assessed at comparable coverage. Whereas recent work provides view-level guarantees for RGB prediction boxes~\cite{chu2026viewstructured}, we target spatially adaptive scalar RGB-MAE bounds with marginal coverage.

To this end, we propose QuantileGS, a post-hoc framework that estimates and calibrates rendering-error uncertainty with a frozen pretrained 3DGS model.
Two nonnegative attributes per Gaussian parameterize ordered lower and upper RGB-MAE quantile fields. Shared across views, these attributes yield view-dependent
predictions through the original $\alpha$-blending weights and fixed background values. Conformal calibration adjusts the upper prediction using the
floored interquantile gap as a local scale~\cite{romano2019conformalized,angelopoulos2023conformal}.
Calibration and test views are excluded from RGB pretraining and quantile fitting. Novel-view bounds require camera parameters but no target RGB.
Under exchangeability of pixel-level calibration and test scores, conditional on fitting, the bounds provide finite-sample marginal
coverage~\cite{lei2018distributionfree}. This guarantee averages over the calibration sample and a new pixel observation, without implying a separate coverage guarantee for each view or pixel location.

Experiments on LF, LLFF, and Mip-NeRF~360 assess error ranking,
empirical coverage, and mean upper bounds. Prior work uses uncertainty for diffusion-based scene refinement~\cite{bose2025uncertaintyaware}, active view selection
(AVS)~\cite{li2026ougs}, Gaussian pruning~\cite{hanson2025pup},
and camera tracking~\cite{tran2026varsplat}. Gaussian representations also support simultaneous localization and mapping (SLAM)~\cite{keetha2024splatam,yan2024gs,wu2025vings} and dynamic driving-scene reconstruction~\cite{xiong2026drivinggaussian}.
We assess downstream utility by using the bounds to weight prediction-variance reduction for AVS, estimate Gaussian deletion costs, and control reference RGB corrections and photometric residual scales for RGB-D SLAM. Fig.~\ref{fig:quantilegs_teaser} illustrates the framework and these applications, including renderings before and after calibrated UQ-guided optimization. Calibration preserves the original RGB rendering. Its coverage guarantee applies to the fixed renderer, while downstream benefits are evaluated empirically.

Our main contributions are summarized as follows:
\begin{itemize}
    \item We introduce ordered Gaussian error-quantile fields
    that learn two attributes per Gaussian while preserving
    a pretrained 3DGS renderer.

    \item We develop scale-normalized conformal calibration
    for spatially adaptive RGB-MAE upper bounds with finite-sample
    marginal coverage under conditional score exchangeability.

    \item We evaluate uncertainty estimation and calibration
    on LF, LLFF, and Mip-NeRF~360, and examine their utility
    in active view selection, Gaussian pruning, and RGB-D SLAM.
\end{itemize}

%% file: sec/2_related_work.tex
\section{Related Work}
\label{sec:related_work}

\subsection{Uncertainty Quantification for View Synthesis}
\label{sec:related_uq}

NVS uncertainty estimates draw on probabilistic modeling,
rendering sensitivity, and multi-view evidence.
S-NeRF~\cite{shen2021stochastic} and
CF-NeRF~\cite{shen2022conditional} model radiance distributions
through variational inference and normalizing flows, respectively.
Bayes' Rays~\cite{goli2024bayes} uses a post-hoc Laplace
approximation.
For 3DGS, approaches include stochastic Gaussian
attributes~\cite{aira2025modeling}, variational multi-scale
representations~\cite{li2024variational}, learned parameter
manifolds~\cite{lyu2024manifold}, and Horseshoe shrinkage
priors~\cite{wu2026horseshoe}.
FisherRF~\cite{jiang2024fisherrf} instead measures sensitivity
through rendering derivatives.

Reconstruction evidence supports error prediction from Gaussian
features~\cite{gottwald2026primu} and uncertainty fitting with
a fixed renderer~\cite{galappaththige2026predictive}.
WarpRF~\cite{safadoust2026warprf} uses cross-view inconsistencies,
while OCSplats~\cite{ling2025ocsplats} uses observation
completeness to refine noise estimates.
Abstract Rendering~\cite{ji2025abstract} certifies image
variations under specified camera and scene perturbations,
rather than errors against observed RGB.
QuantileGS learns ordered rendering-error quantiles and
calibrates their upper prediction while preserving the RGB model.

\subsection{Conformal Prediction and Error Calibration}
\label{sec:related_conformal}

Split conformal prediction provides finite-sample marginal
coverage under exchangeability~\cite{lei2018distributionfree}.
Residual normalization~\cite{lei2018distributionfree} and
conformalized quantile regression~\cite{romano2019conformalized}
adapt interval widths, while score optimization targets
efficiency~\cite{xie2024boosted}. Conformal risk control extends calibration to expected bounded monotone losses~\cite{angelopoulos2024conformal}, with applications to confidence masks for generative super-resolution~\cite{adame2025superresolution}. Further work studies guarantees for specified subgroups and covariate shifts~\cite{gibbs2025conditional}, and calibration under distribution shift using optimal transport and unlabeled target data~\cite{correia2025transport}.

For NVS, Meta-Calibrator~\cite{amini2024instant} learns scene-specific calibration curves across scenes. VSCP~\cite{chu2026viewstructured} constructs RGB prediction
boxes that cover a prescribed fraction of pixels with a specified probability under view-level exchangeability. QuantileGS calibrates a scalar RGB-MAE bound using the predicted interquantile gap for spatial normalization. Its guarantee is marginal under exchangeability of pixel-level calibration and test scores, rather than conditional at every pixel or view.

\subsection{Downstream Applications of Radiance Fields}
\label{sec:related_downstream}

View acquisition methods use predicted uncertainty reduction~\cite{pan2022activenerf}, information-based criteria~\cite{jiang2024fisherrf,wilson2025pop},
or anisotropic visibility~\cite{xue2026uncertainty}.
Gaussian compression uses global significance~\cite{fan2024lightgaussian}, reconstruction-error sensitivity~\cite{hanson2025pup}, or pruning integrated with efficient rasterization~\cite{hanson2025speedysplat}.

Gaussian representations also support RGB-D tracking and mapping~\cite{keetha2024splatam,yugay2023gaussianslam}.
SEGS-SLAM~\cite{wen2025segsslam} incorporates structural cues and appearance embeddings, while
WildGS-SLAM~\cite{zheng2025wildgs} and
VarSplat~\cite{tran2026varsplat} incorporate learned uncertainty into tracking.
QuantileGS uses calibrated error bounds to weight prediction-variance reduction for acquisition, rank pruning risk, and scale photometric tracking residuals.

%% file: sec/3_preliminaries.tex
\section{Preliminaries}
\label{sec:preliminaries}

\subsection{3D Gaussian Splatting}
\label{sec:prelim_3dgs}

3DGS~\cite{kerbl20233d} represents a scene by Gaussian primitives $\mathcal G$, with pretrained parameters $\boldsymbol\theta$ encoding their centers, scales, rotations, opacities, and view-dependent color coefficients.

For a camera ray $\boldsymbol x$, let $\alpha_j(\boldsymbol x)\in[0,1]$ be the effective opacity of Gaussian $j$, incorporating its opacity and projected footprint. Front-to-back compositing uses
\begin{equation}
\begin{aligned}
    T_j(\boldsymbol x)
    &=\prod_{\ell\prec_{\boldsymbol x}j}
      \bigl(1-\alpha_\ell(\boldsymbol x)\bigr),\\
    w_j(\boldsymbol x)
    &=T_j(\boldsymbol x)\alpha_j(\boldsymbol x),
\end{aligned}
\label{eq:3dgs_weight}
\end{equation}
where $\ell\prec_{\boldsymbol x}j$ denotes depth ordering along the ray, $T_j$ is transmittance, and $w_j$ is the blending weight. The accumulated opacity and residual transmittance are
\begin{equation}
    A(\boldsymbol x)=\sum_{j\in\mathcal G}w_j(\boldsymbol x),
    \qquad
    r(\boldsymbol x)=1-A(\boldsymbol x).
\label{eq:3dgs_opacity}
\end{equation}
The Gaussian and background weights are therefore
nonnegative and sum to one.

For Gaussian colors $\boldsymbol c_j(\boldsymbol d)$ at viewing direction $\boldsymbol d$ and a fixed RGB background $\boldsymbol b\in[0,1]^3$, the composited color is
\begin{equation}
    \boldsymbol C(\boldsymbol x)
    =\sum_{j\in\mathcal G}w_j(\boldsymbol x)
      \boldsymbol c_j(\boldsymbol d)
      +r(\boldsymbol x)\boldsymbol b.
\label{eq:3dgs_rendering}
\end{equation}

The RGB prediction at pixel $i$ is
\[
    \hat{\boldsymbol y}_i
    =f_{\boldsymbol\theta}(\boldsymbol x_i)
    =\operatorname{clip}_{[0,1]^3}
      \bigl(\boldsymbol C(\boldsymbol x_i)\bigr).
\]

Freezing $\boldsymbol\theta$ fixes the weights for each ray. QuantileGS reuses them to composite learned Gaussian error attributes with fixed scalar background values.

\subsection{Problem Formulation}
\label{sec:problem_setting}

Let $X$ be a random pixel ray and $\boldsymbol Y\in[0,1]^3$ its ground-truth RGB.
For the fixed renderer, the RGB-MAE is $E=\|f_{\boldsymbol\theta}(X)-\boldsymbol Y\|_1/3\in[0,1]$, with observed values $e_i$ defined in
Eq.~\eqref{eq:rendering_error}. Its conditional quantile at level $\nu\in(0,1)$ is
\begin{equation}
    q_\nu(\boldsymbol x)
    =\inf\bigl\{t:
      \mathbb P(E\leq t\mid X=\boldsymbol x)\geq\nu
      \bigr\}.
\label{eq:error_quantile}
\end{equation}

For a novel view without target RGB, we seek a pixelwise upper error bound $\hat u_\tau$ satisfying
\begin{equation}
    \mathbb P\!\left\{
        E_{\mathrm{new}}\leq\hat u_\tau(X_{\mathrm{new}})
    \right\}\geq\tau,
\label{eq:coverage_goal}
\end{equation}
where $\tau\in(0,1)$ is the target coverage. The probability is marginal over the calibration sample and a new pixel observation, with fitting information
held fixed.

%% file: sec/4_quantile.tex
\section{QuantileGS}
\label{sec:quantilegs}

\begin{figure*}[t]
    \centering
    \includegraphics[width=\textwidth]{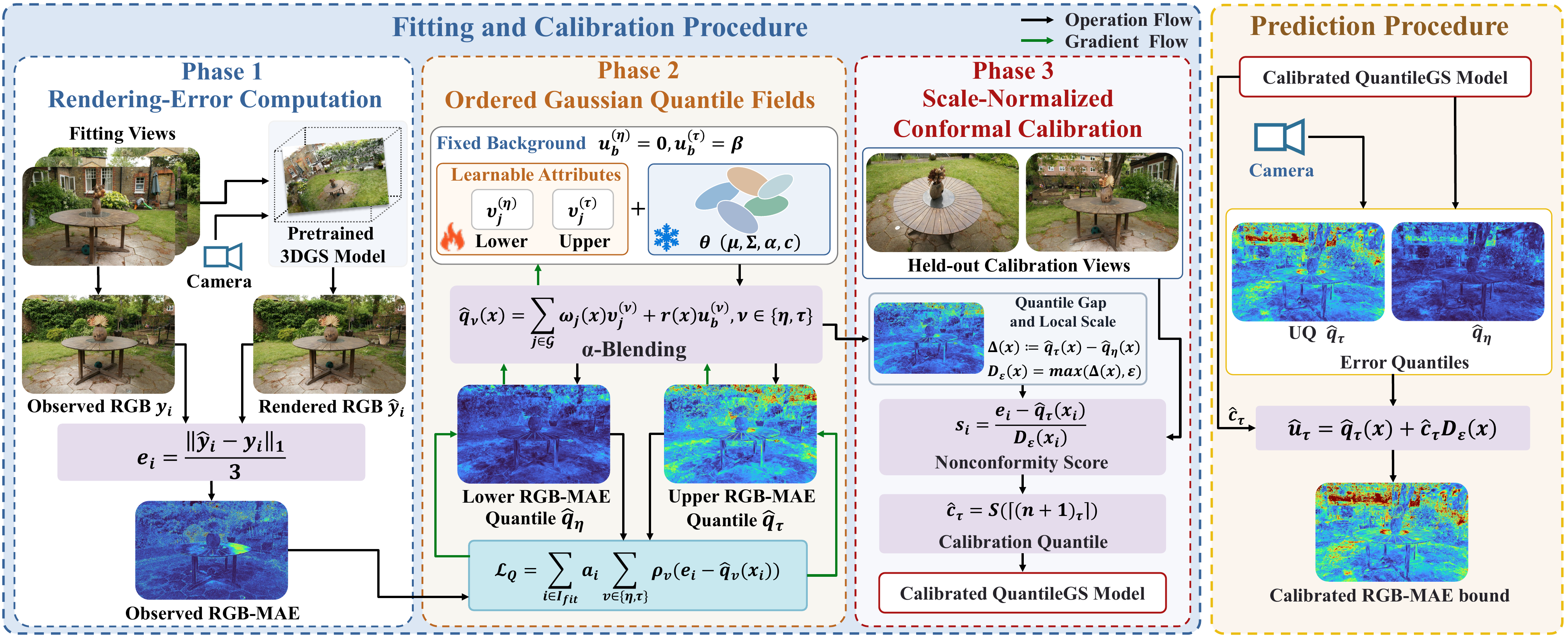}
    \caption{
        \textbf{QuantileGS overview.}
        Phase~1 computes RGB-MAE on $\mathcal V_{\mathrm{fit}}$
        using a frozen pretrained 3DGS model.
        Phase~2 learns ordered Gaussian error quantiles with
        fixed blending weights and background contributions.
        Phase~3 calibrates the upper prediction on
        $\mathcal V_{\mathrm{cal}}$, using the interquantile
        gap with a positive floor as the local scale.
        The prediction procedure renders the learned quantiles for a new view and applies
        a locally scaled conformal correction to produce an RGB-MAE
        upper-bound map.
    }
    \label{fig:quantilegs}
\end{figure*}

QuantileGS constructs calibrated pixel-wise RGB-MAE bounds
through the three phases in Fig.~\ref{fig:quantilegs}.
We partition views into disjoint \emph{fitting views}
$\mathcal V_{\mathrm{fit}}$, \emph{calibration views}
$\mathcal V_{\mathrm{cal}}$, and \emph{test views}
$\mathcal V_{\mathrm{test}}$, with corresponding pixel-index
sets $\mathcal I_{\mathrm{fit}}$, $\mathcal I_{\mathrm{cal}}$,
and $\mathcal I_{\mathrm{test}}$.
We use $\mathcal V_{\mathrm{fit}}$ for RGB pretraining and
quantile fitting, $\mathcal V_{\mathrm{cal}}$ only for
calibration, and $\mathcal V_{\mathrm{test}}$ only for evaluation.

\subsection{Rendering-Error Computation}
\label{sec:rendering_error_computation}

We start from a 3DGS model pretrained on $\mathcal V_{\mathrm{fit}}$.
All original parameters $\boldsymbol\theta$ remain frozen throughout QuantileGS. For a pixel ray $\boldsymbol x_i$ specified by its camera parameters, let $\boldsymbol y_i\in[0,1]^3$ be the observed RGB and $\hat{\boldsymbol y}_i=f_{\boldsymbol\theta} (\boldsymbol x_i)\in[0,1]^3$ the clipped rendering. The fixed fitting targets are
\begin{equation}
    e_i=\frac{\|\hat{\boldsymbol y}_i-\boldsymbol y_i\|_1}{3}
    \in[0,1], i\in\mathcal I_{\mathrm{fit}}.
\label{eq:rendering_error}
\end{equation}

\subsection{Ordered Gaussian Quantile Fields}
\label{sec:gaussian_quantile_field}

For each Gaussian $j\in\mathcal G$, we learn attributes $v_j^{(\eta)}$ and $v_j^{(\tau)}$ for lower and upper error quantiles at levels $0<\eta<\tau<1$. We use $\eta=0.5$, and $\tau$ is the target coverage level.

For the RGB background $\boldsymbol b\in[0,1]^3$, define
\begin{equation}
    d_i=\frac{\|\boldsymbol y_i-\boldsymbol b\|_1}{3},
     i\in\mathcal I_{\mathrm{fit}}.
\label{eq:background_error_samples}
\end{equation}

Sort these $N_{\mathrm{fit}}=|\mathcal I_{\mathrm{fit}}|$
values as $d_{(1)}\leq\cdots\leq d_{(N_{\mathrm{fit}})}$ and set
\begin{equation}
    \beta=d_{(\lceil N_{\mathrm{fit}}\tau\rceil)},
    \quad u_b^{(\eta)}=0, u_b^{(\tau)}=\beta.
\label{eq:fixed_background_quantiles}
\end{equation}

Thus, $\beta$ is an empirical upper quantile of the observed-color-to-background error. We compute it once from fitting colors and store it as a fixed buffer without gradients.

Let $w_j(\boldsymbol x)\geq0$ be the frozen
$\alpha$-blending weights and
$r(\boldsymbol x)=1-\sum_jw_j(\boldsymbol x)\geq0$
the residual transmittance. For $\nu\in\{\eta,\tau\}$,
\begin{equation}
    \hat q_\nu(\boldsymbol x)
    =\sum_{j\in\mathcal G}w_j(\boldsymbol x)v_j^{(\nu)}
     +r(\boldsymbol x)u_b^{(\nu)}.
\label{eq:gaussian_quantile_field}
\end{equation}

Phase~2 of Fig.~\ref{fig:quantilegs} composites Gaussian-specific scalar attributes into pixel-wise error-quantile predictions. The index $j$ is summed out in $\hat q_\nu(\boldsymbol x)$, as in the RGB rendering $\boldsymbol C(\boldsymbol x)$,
while each $v_j^{(\nu)}$ remains specific to Gaussian $j$. Unlike the directional colors $\boldsymbol c_j(\boldsymbol d)$, these scalar attributes are shared across views, with
view dependence entering through the blending weights $w_j(\boldsymbol x)$ and residual transmittance $r(\boldsymbol x)$. The attributes are learned jointly from pixel-wise RGB-MAE targets, without requiring separate error targets for individual Gaussians.

The lower field contains only Gaussian contributions, and the upper field includes the fixed offset $r(\boldsymbol x)\beta$. We impose the ordering constraints
\begin{equation}
    v_j^{(\eta)}\geq0,
    v_j^{(\tau)}-v_j^{(\eta)}\geq0,
     j\in\mathcal G,
\label{eq:ordered_constraints}
\end{equation}
using $v_j^{(\eta)}=\ell_j$ and
$v_j^{(\tau)}=\ell_j+h_j$ with $\ell_j,h_j\geq0$.
Since $\beta\geq0$, the composited gap satisfies
\begin{equation}
\begin{aligned}
    \Delta(\boldsymbol x)
    &:=\hat q_\tau(\boldsymbol x)-\hat q_\eta(\boldsymbol x)\\
    &=\sum_{j\in\mathcal G}w_j(\boldsymbol x)h_j
      +r(\boldsymbol x)\beta\geq0.
\end{aligned}
\label{eq:quantile_gap}
\end{equation}

The fields cannot cross, but their gap can vanish, motivating the positive floor in the calibration scale $D_\varepsilon$ below.

With the pinball loss
\begin{equation}
    \rho_\nu(z)=\max\{\nu z,(\nu-1)z\},
\label{eq:pinball_loss}
\end{equation}
we jointly fit both fields by minimizing
\begin{equation}
    \mathcal L_{\mathrm{UQ}}
    =\sum_{i\in\mathcal I_{\mathrm{fit}}}a_i
      \sum_{\nu\in\{\eta,\tau\}}
      \rho_\nu\!\left(e_i-\hat q_\nu(\boldsymbol x_i)\right),
\label{eq:quantile_objective}
\end{equation}
where $a_i\geq0$ are fixed weights summing to one.
We project $\ell_j,h_j$ onto $[0,\infty)$ after each
first-order update. Gradients pass through the fixed blending
operation to these attributes; the RGB parameters
$\boldsymbol\theta$, weights $w_j$, and background statistic
$\beta$ remain fixed.

To establish convexity in attribute space, stack the Gaussian attributes and pixel-$i$ weights into $\boldsymbol v^{(\nu)}$ and $\boldsymbol w_i$.
With fixed offsets $b_{i\eta}=0$ and
$b_{i\tau}=r(\boldsymbol x_i)\beta$, the predictions are affine:
$\hat q_\nu(\boldsymbol x_i)
=\boldsymbol w_i^\top\boldsymbol v^{(\nu)}+b_{i\nu}$.
Each loss term is a maximum of affine functions, so Eq.~\eqref{eq:quantile_objective} is convex under the linear ordering constraints. Introducing auxiliary variables
$\xi_{i\nu}$ gives the equivalent linear program
\begin{equation}
\begin{aligned}
    \min_{\substack{\boldsymbol v^{(\eta)},\,\boldsymbol v^{(\tau)},\\
                    \boldsymbol\xi}}\quad&
    \sum_{i\in\mathcal I_{\mathrm{fit}}}
    \sum_{\nu\in\{\eta,\tau\}}a_i\xi_{i\nu}\\
    \mathrm{s.t.}\quad&
    \xi_{i\nu}\geq\nu
       (e_i-\boldsymbol w_i^\top\boldsymbol v^{(\nu)}-b_{i\nu}),\\
    &\xi_{i\nu}\geq(\nu-1)
       (e_i-\boldsymbol w_i^\top\boldsymbol v^{(\nu)}-b_{i\nu}),\\
    &\boldsymbol v^{(\eta)}\geq\boldsymbol 0,\\
    &\boldsymbol v^{(\tau)}-\boldsymbol v^{(\eta)}
      \geq\boldsymbol 0.
\end{aligned}
\label{eq:quantile_linear_program}
\end{equation}

The first two constraints apply to every fitting pixel and both quantile levels, with vector inequalities taken componentwise. Background offsets are fixed rather than optimized. In practice, we optimize the pinball objective directly.

\subsection{Scale-Normalized Conformal Calibration}
\label{sec:conformal_calibration}

We freeze the fitted fields and use $n=|\mathcal I_{\mathrm{cal}}|$ pixel observations from $\mathcal V_{\mathrm{cal}}$ to calibrate the upper prediction. Fix $\varepsilon>0$ and $\tau$ before calibration, and define
\begin{equation}
    D_\varepsilon(\boldsymbol x)
    =\max\{\Delta(\boldsymbol x),\varepsilon\}.
\label{eq:floored_quantile_scale}
\end{equation}

The interquantile gap measures the predicted spread between the
median and the target upper quantile. It allows the calibration correction to vary spatially, while the positive floor prevents division by a vanishing gap.

Reindexing the calibration pixels by $i=1,\ldots,n$, compute
their RGB-MAE and normalized scores
\begin{equation}
    s_i=\frac{e_i-\hat q_\tau(\boldsymbol x_i)}
                  {D_\varepsilon(\boldsymbol x_i)}.
\label{eq:normalized_score}
\end{equation}

Positive scores indicate that the observed error exceeds the upper
prediction, whereas negative scores indicate that it falls below it.
Normalization expresses these deviations relative to the floored
local scale, rather than their absolute RGB-MAE magnitude.

Sort the scores as $s_{(1)}\leq\cdots\leq s_{(n)}$,
append $s_{(n+1)}=+\infty$, and set
\begin{equation}
    \hat c_\tau=s_{(k_\tau)},
     k_\tau=\lceil(n+1)\tau\rceil.
\label{eq:conformal_coefficient}
\end{equation}

The scalar correction adapts to each ray through the same scale:
\begin{equation}
    \hat u_\tau(\boldsymbol x)
    =\hat q_\tau(\boldsymbol x)
     +\hat c_\tau D_\varepsilon(\boldsymbol x).
\label{eq:final_quantilegs}
\end{equation}

Phase~3 of Fig.~\ref{fig:quantilegs} illustrates how a single
calibration coefficient produces spatially varying corrections
through $D_\varepsilon(\boldsymbol x)$. A positive $\hat c_\tau$ increases the upper predictions, whereas a negative value reduces them. Both the fitted quantile fields and the pretrained RGB renderer remain frozen. Calibration therefore changes the reported error bounds without changing the RGB predictions or their realized errors. Let $\mathcal F_{\mathrm{fit}}$ contain all fitting information, including RGB pretraining and optimization randomness. The following theorem explains the gap normalization.

\begin{qgstheorem}[Joint local-scale calibration]
\label{thm:quantilegs_coverage}
Condition on $\mathcal F_{\mathrm{fit}}$.
For the $n$ calibration observations and one new test
observation, suppose
$E_i=\mu(X_i)+\sigma(X_i)Z_i$ and
$\hat q_\nu(\boldsymbol x)
=\mu(\boldsymbol x)+\sigma(\boldsymbol x)\hat z_\nu$,
$\nu\in\{\eta,\tau\}$, with fixed functions $\mu,\sigma$,
$\sigma>0$, and finite fitting-dependent constants
$\hat z_\eta<\hat z_\tau$.
Assume the standardized residuals are i.i.d.\ and their
vector is independent of the entire ray vector.
If $\Delta(X_i)\geq\varepsilon$ almost surely for
$i=1,\ldots,n+1$, the normalized scores are conditionally
i.i.d.\ and
\begin{equation}
    \mathbb P\!\left\{
      E_{\mathrm{new}}\leq\hat u_\tau(X_{\mathrm{new}})
      \mid\mathcal F_{\mathrm{fit}}\right\}\geq\tau.
\label{eq:quantilegs_coverage}
\end{equation}
\end{qgstheorem}

The proof is in Appendix~\ref{sec:proof_quantilegs_coverage}.
Coverage is marginal, and view separation alone does not establish the assumptions. For an active floor, exchangeability of the actual floored scores is sufficient for the same guarantee. This marginal guarantee does not imply simultaneous coverage of all pixels in a view or a separate coverage guarantee for each view.

\subsection{Prediction Procedure}
\label{sec:prediction_procedure}

After calibration, the RGB model, Gaussian error attributes, background values, and calibration parameters remain fixed. Given a new camera pose and intrinsics, we render the RGB image with $f_{\boldsymbol\theta}$ and the two quantile fields
using Eq.~\eqref{eq:gaussian_quantile_field}. Their floored gap, defined in
Eq.~\eqref{eq:floored_quantile_scale}, scales the correction in Eq.~\eqref{eq:final_quantilegs}, yielding $\hat u_\tau(\boldsymbol x)$.
We return
\begin{equation}
    u(\boldsymbol x)=\operatorname{clip}_{[0,1]}
        \bigl(\hat u_\tau(\boldsymbol x)\bigr).
\label{eq:clipped_rgb_mae_bound}
\end{equation}

Clipping preserves the coverage guarantee because RGB-MAE
lies in $[0,1]$. Prediction requires no target RGB and returns the unchanged
RGB rendering together with a spatially aligned RGB-MAE upper-bound map.

\subsection{Downstream Applications}
\label{sec:downstream_applications}

We use calibrated RGB-MAE bounds $u(\boldsymbol x)$ to guide three downstream applications. For \emph{active view selection}, the bounds weight the
reduction of a local prediction-variance proxy, favoring views informative about regions with larger bounds. For \emph{Gaussian pruning}, they combine with estimated
deletion perturbations to assess deletion costs while accounting for unknown rendering residuals. For \emph{RGB-D SLAM}, they limit reference-color corrections
toward observed fitting RGB and set robust scales for photometric residuals during pose estimation. The coverage guarantee applies to rendering errors of the fixed model, while downstream benefits are evaluated empirically. Application-specific objectives, observation splits, and implementation details are provided in
Appendix~\ref{sec:supp_downstream_applications}.

%% file: sec/5_experiments.tex
\section{Experiments}
\label{sec:experiments}

\subsection{Uncertainty Quantification in NVS}
\label{sec:uq_nvs}

\begin{table*}[!t]
    \centering
    \caption{
        Uncertainty quantification, empirical coverage, and novel-view
        synthesis results on LLFF, Mip-NeRF~360, and LF.
        The \colorbox{bestcolor}{best},
        \colorbox{secondcolor}{second-best}, and
        \colorbox{thirdcolor}{third-best} results for AUSE and Pearson
        correlation are highlighted, with the best also in bold.
        Coverage is the mean empirical coverage across scenes,
        with a nominal coverage level of $90\%$ ($\alpha = 0.1$).
        It is reported for both QuantileGS variants;
        \textemdash{} denotes baselines without conformal calibration,
        for which coverage is not reported.
    }
    \label{tab:uq_llff_mip360}
    \small
    \setlength{\tabcolsep}{4pt}
    \renewcommand{\arraystretch}{1.12}
    \begin{tabular*}{\textwidth}{@{\extracolsep{\fill}}llcccccc@{}}
        \toprule
        Dataset & Method
        & AUSE $\downarrow$ & Pears. $\uparrow$
        & Coverage (\%)
        & PSNR $\uparrow$ & SSIM $\uparrow$ & LPIPS $\downarrow$ \\
        \midrule

        \multirow{6}{*}{LLFF}
        & Variational-3DGS
        & 0.381 & 0.269 & \textemdash
        & 26.179 & 0.847 & 0.143 \\
        & FisherRF
        & 0.420 & -0.059 & \textemdash
        & 24.880 & 0.824 & 0.156 \\
        & PRIMU
        & \third{0.289} & \third{0.364} & \textemdash
        & 24.508 & 0.820 & 0.163 \\
        & Manifold
        & 0.436 & 0.186 & \textemdash
        & 25.311 & 0.826 & 0.151 \\
        & QuantileGS-U (ours)
        & \qgsSecond{0.250} & \best{\textbf{0.428}} & 69.6
        & \multirow{2}{*}{24.407}
        & \multirow{2}{*}{0.796}
        & \multirow{2}{*}{0.173} \\
        & QuantileGS-C (ours)
        & \best{\textbf{0.249}} & \qgsSecond{0.421} & 89.6
        & & & \\
        \midrule

        \multirow{6}{*}{Mip-NeRF~360}
        & Variational-3DGS
        & 0.514 & 0.143 & \textemdash
        & 27.350 & 0.812 & 0.225 \\
        & FisherRF
        & 0.476 & -0.050 & \textemdash
        & 27.644 & 0.823 & 0.206 \\
        & PRIMU
        & \third{0.389} & 0.214 & \textemdash
        & 27.686 & 0.825 & 0.201 \\
        & Manifold
        & 0.444 & \third{0.221} & \textemdash
        & 27.696 & 0.823 & 0.207 \\
        & QuantileGS-U (ours)
        & \best{\textbf{0.315}} & \best{\textbf{0.371}} & 80.1
        & \multirow{2}{*}{27.165}
        & \multirow{2}{*}{0.810}
        & \multirow{2}{*}{0.214} \\
        & QuantileGS-C (ours)
        & \qgsSecond{0.316} & \qgsSecond{0.368} & 92.1
        & & & \\
        \midrule

        \multirow{6}{*}{LF}
        & Variational-3DGS
        & \third{0.333} & 0.351 & \textemdash
        & 18.639 & 0.673 & 0.319 \\
        & FisherRF
        & 0.542 & -0.070 & \textemdash
        & 26.722 & 0.899 & 0.140 \\
        & PRIMU
        & \qgsSecond{0.321} & \third{0.353} & \textemdash
        & 18.487 & 0.618 & 0.350 \\
        & Manifold
        & 0.434 & 0.173 & \textemdash
        & 25.924 & 0.877 & 0.157 \\
        & QuantileGS-U (ours)
        & \best{\textbf{0.275}} & \qgsSecond{0.392} & 96.3
        & \multirow{2}{*}{28.967}
        & \multirow{2}{*}{0.935}
        & \multirow{2}{*}{0.072} \\
        & QuantileGS-C (ours)
        & \best{\textbf{0.275}} & \best{\textbf{0.393}} & 96.4
        & & & \\

        \bottomrule
    \end{tabular*}
\end{table*}

\noindent\textbf{Datasets and protocol.}
We evaluate on LF, LLFF, and Mip-NeRF~360.
LF~\cite{yucer2016efficient} contains four scenes:
africa, basket, statue, and torch.
Following~\cite{li2024variational}, we retain 4--5 reconstruction views and four test views per scene. We select eight additional views using camera poses and split them into four for fitting directional correction and four for gap-based calibration, none of which are used for reconstruction or testing. LLFF~\cite{mildenhall2019local} contains eight forward-facing scenes,
with every eighth image held out for testing and images downscaled
by $4\times$.
Mip-NeRF~360~\cite{barron2022mip} contains nine scenes.
Following~\cite{kerbl20233d}, we hold out every eighth image and downscale indoor and outdoor images by $2\times$ and $4\times$, respectively. 
QuantileGS uses disjoint fitting and calibration sets, with the underlying 3DGS model frozen throughout UQ fitting and calibration. All experiments use an NVIDIA RTX 3090 with 48 GB memory.

\noindent\textbf{Baselines and metrics.}
We compare against Variational-3DGS~\cite{li2024variational},
FisherRF~\cite{jiang2024fisherrf},
PRIMU~\cite{gottwald2026primu}, and
Manifold~\cite{lyu2024manifold}, using their official implementations.
We report PSNR, SSIM, and LPIPS for rendering, and
MAE-based AUSE~\cite{ilg2018uncertainty} and Pearson correlation with RGB-MAE for uncertainty quality. QuantileGS-U and QuantileGS-C denote predictions before and
after calibration and share the same frozen RGB renderer.
For each scene, let $N_{\mathrm{test}}=|\mathcal I_{\mathrm{test}}|$. For QuantileGS-C, empirical coverage and mean bound are
\begin{equation}
\begin{aligned}
    \widehat{\mathrm{Cov}}_\tau
    &= \frac{1}{N_{\mathrm{test}}}
       \sum_{i\in\mathcal I_{\mathrm{test}}}
       \mathds{1}\!\left\{
           e_i \leq u(\boldsymbol x_i)
       \right\},\\
    \overline{u}
    &= \frac{1}{N_{\mathrm{test}}}
       \sum_{i\in\mathcal I_{\mathrm{test}}}
       u(\boldsymbol x_i),
\end{aligned}
\label{eq:empirical_coverage_mean_bound}
\end{equation}
where $e_i$ is the observed RGB-MAE at test pixel $i$. For QuantileGS-U, we replace $u(\boldsymbol x_i)$ with the unclipped upper quantile $\hat q_\tau(\boldsymbol x_i)$
in both expressions.

Coverage measures the fraction of test errors contained by their predicted bounds. At comparable coverage, a smaller mean bound indicates tighter uncertainty estimates.
AUSE assesses error ranking, whereas Pearson correlation measures linear association with RGB-MAE. Both coverage and mean bound pool all test pixels within each scene, and dataset-level results average scenes equally.

\begin{figure*}[!t]
    \centering
    \includegraphics[width=\textwidth]{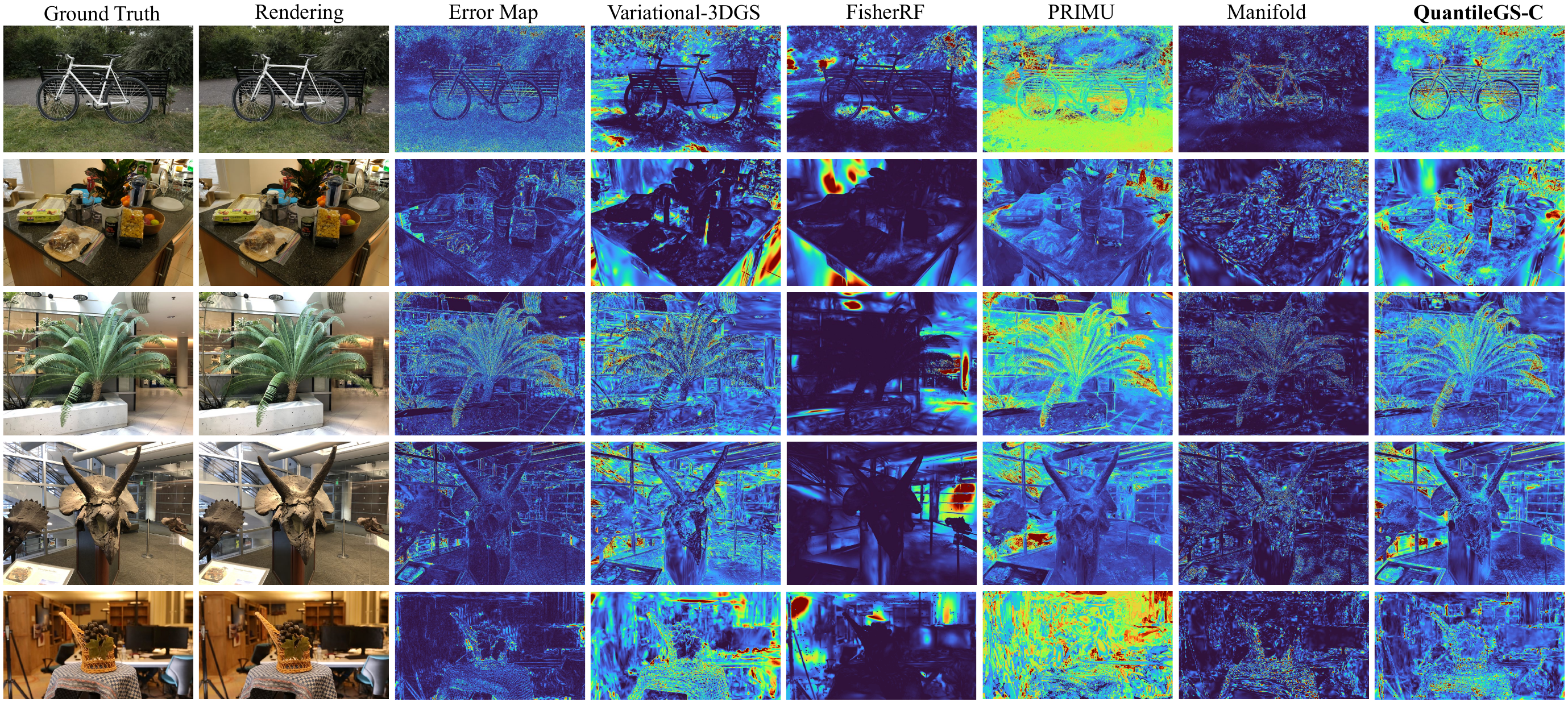}
    \caption{
    Qualitative comparison of Variational-3DGS~\cite{li2024variational},
    FisherRF~\cite{jiang2024fisherrf}, PRIMU~\cite{gottwald2026primu}, Manifold~\cite{lyu2024manifold}, and QuantileGS-C (Ours).
    Each scene shows uncertainty maps.
    Scenes from top to bottom: Mip-NeRF~360 bicycle and counter, LLFF fern and horns, and LF basket.
    }
    \label{fig:qualitative_uq}
\end{figure*}

\noindent\textbf{Results.}
Tables~\ref{tab:uq_llff_mip360} show that QuantileGS achieves lower mean AUSE and higher Pearson correlation than all baselines on all three datasets.
Calibration leaves these metrics broadly unchanged. Fig.~\ref{fig:qualitative_uq} compares the spatial correspondence between uncertainty estimates and rendering errors. Calibration raises mean coverage from $69.6\%$ to $89.6\%$ on LLFF and from $80.1\%$ to $92.1\%$ on Mip-NeRF~360, while AUSE and correlation change little. The $96.4\%$ coverage on LF indicates conservative bounds in this setting. More experiments are provided in Appendix~\ref{sec:supp_uq_results}.

\subsection{Downstream Applications}

\subsubsection{Active View Selection}
\label{sec:avs}

\noindent\textbf{Setting and baselines.}
We evaluate on nine Mip-NeRF~360 scenes with a budget of 20 reconstruction views. Every eighth view is held out for testing. Indoor and outdoor images are downscaled by $2\times$ and $4\times$, respectively. We compare against Random, FisherRF, Manifold, and PRIMU. All methods use the same four initial views, comprising one fixed view and three views selected sequentially by farthest-point sampling.
QuantileGS uses the acquisition rule in
Sec.~\ref{sec:active_view_selection}, adding one view every $100n$ optimization steps, where $n$ is the current number of selected views. After each acquisition, 3DGS restarts from the same initialization; the final model is evaluated at global
iteration 30,000, without additional reconstruction.
RGB training uses random backgrounds, while UQ and evaluation use fixed black backgrounds.
Reserved UQ fitting and calibration views are additional observations, excluded from our reconstruction set and candidate pool, and baseline candidate pools remain unchanged.
Manifold retains its VI\_MC acquisition and 2,000-step per-round training, followed by a separate 30,000-step vanilla 3DGS reconstruction using its selected 20 views. Random results are averaged over seeds 0, 1, and 2. All experiments use an NVIDIA RTX 3090 with 24 GB memory.

\begin{table}[!tb]
    \centering
    \caption{
        AVS on Mip-NeRF~360: nine-scene means with
        20 reconstruction views.
        The \colorbox{bestcolor}{best},
        \colorbox{secondcolor}{second-best}, and
        \colorbox{thirdcolor}{third-best} results are highlighted,
        with the best also in bold.
        Ties at the displayed precision share a rank.
    }
    \label{tab:avs_results}
    \small
    \setlength{\tabcolsep}{4pt}
    \renewcommand{\arraystretch}{1.12}
    \begin{tabular*}{\columnwidth}{@{\extracolsep{\fill}}lccc@{}}
        \toprule
        Method
        & PSNR $\uparrow$
        & SSIM $\uparrow$
        & LPIPS $\downarrow$ \\
        \midrule
        Random
        & 19.361 & 0.584 & 0.364 \\
        FisherRF
        & 20.316 & 0.597 & \third{0.355} \\
        Manifold
        & \third{20.543} & \third{0.609} & \qgsSecond{0.338} \\
        PRIMU
        & \qgsSecond{20.846} & \best{\textbf{0.636}}
        & \best{\textbf{0.328}} \\
        QuantileGS (Ours)
        & \best{\textbf{21.090}} & \qgsSecond{0.635}
        & \best{\textbf{0.328}} \\
        \bottomrule
    \end{tabular*}
\end{table}

\noindent\textbf{Results.}
Table~\ref{tab:avs_results} shows that QuantileGS achieves the
highest PSNR, exceeding PRIMU and Manifold by $0.244$ and $0.547$\,dB, respectively.
Its SSIM is $0.001$ below PRIMU, while LPIPS ties for best at the reported precision.
These results compare the complete AVS pipelines under the settings above. Relative to Random, QuantileGS improves PSNR by $1.729$\,dB
and reduces LPIPS by $0.036$, supporting the practical utility of the evaluated acquisition pipeline. 

\subsubsection{Gaussian Pruning}
\label{sec:pruning_experiments}

\noindent\textbf{Setting and baselines.}
We evaluate the strategy in Sec.~\ref{sec:gaussian_pruning}
on nine Mip-NeRF~360 scenes against
PUP~3D-GS~\cite{hanson2025pup} and
LightGaussian~\cite{fan2024lightgaussian}, using identical 30k-iteration checkpoints, resolutions, and test views. 
We reuse the saved calibration splits, containing 21-54 views per scene, excluded from base-model and UQ fitting. QuantileGS uses projected parameters with Adam, $\tau=0.9$, $\alpha=0.1$, gap normalization, and two probes per fitting view.
All methods retain 20\% and then 10\% of the original
Gaussians. Our method uses two progressive pruning stages per round, with 2.5k refinement updates per stage and no densification. Scoring uses images downscaled by a further factor of four. Refinement uses all non-test training views with UQ frozen.
LightGaussian uses volume-adjusted importance pruning and recovery without SH distillation or quantization. All experiments use an NVIDIA RTX 3090 with 24 GB memory.

\begin{table}[!tb]
    \centering
    \caption{
        Pruning on Mip-NeRF~360: nine-scene means, seed 0.
        Within each retention level, the
        \colorbox{bestcolor}{best} and
        \colorbox{secondcolor}{second-best} results are highlighted,
        with the best also in bold.
        Rendering metrics are computed from floating-point images.
    }
    \label{tab:pruning_results}
    \small
    \setlength{\tabcolsep}{3pt}
    \renewcommand{\arraystretch}{1.12}
    \begin{tabular*}{\columnwidth}{@{\extracolsep{\fill}}lcccc@{}}
        \toprule
        Method & Kept
        & PSNR $\uparrow$ & SSIM $\uparrow$ & LPIPS $\downarrow$ \\
        \midrule
        Unpruned 3DGS
        & 100\% & 27.170 & 0.810 & 0.215 \\
        \midrule
        LightGaussian
        & 20\% & 26.814 & \qgsSecond{0.794} & \qgsSecond{0.257} \\
        PUP 3D-GS
        & 20\% & \qgsSecond{27.037}
        & \best{\textbf{0.801}} & \best{\textbf{0.242}} \\
        QuantileGS (Ours)
        & 20\% & \best{\textbf{27.090}} & 0.784 & 0.274 \\
        \midrule
        LightGaussian
        & 10\% & 26.328 & \qgsSecond{0.770} & \qgsSecond{0.300} \\
        PUP 3D-GS
        & 10\% & \qgsSecond{26.579}
        & \best{\textbf{0.785}} & \best{\textbf{0.274}} \\
        QuantileGS (Ours)
        & 10\% & \best{\textbf{26.849}} & 0.769 & 0.303 \\
        \bottomrule
    \end{tabular*}
\end{table}

\noindent\textbf{Results.}
Table~\ref{tab:pruning_results} shows that QuantileGS achieves the highest PSNR at both retention levels, although both baselines retain better SSIM and LPIPS.
At 20\% and 10\% retention, the mean Gaussian count decreases from 2.44 million to 0.488 million and 0.244 million. RGB-only PLY storage falls from 577.0 to 115.4 and 57.7\, MiB, while RGB rendering throughput increases from 112.7 to 284.4 and 348.4 FPS, respectively.
Empirical coverage increases from 92.14\% to 96.44\% and 96.72\%
at 20\% and 10\% retention, respectively, while mean error bounds
widen from 0.0929 to 0.1125 and 0.1159. On the evaluated test pixels, no originally covered pixel becomes uncovered under the transported bounds after pruning and
refinement, within a numerical tolerance of $10^{-6}$. The increased coverage reflects more conservative transported bounds, whose evaluation additionally requires the original
model and its quantile fields. These additional costs are excluded from the reported storage and rendering throughput.
At $10\%$ retention, the PSNR drop from the unpruned model
is $0.321$\,dB for QuantileGS, compared with $0.591$\,dB for
PUP~3D-GS and $0.842$\,dB for LightGaussian.
The results show a trade-off between PSNR preservation and
structural/perceptual similarity. More experiments are provided in Appendix~\ref{sec:supp_pruning_results}.

\subsubsection{Simultaneous Localization and Mapping}
\label{sec:slam_experiments}

\noindent\textbf{Setting and baselines.}
We evaluate on eight Replica RGB-D sequences (room0-2 and office0-4), using all 2,000 frames at
$1200\!\times\!680$ resolution, stride 1, and seed 0. All experiments use an NVIDIA RTX 3090 with 24\, GB memory. We reproduce Gaussian-SLAM~\cite{yugay2023gaussianslam},
VarSplat~\cite{tran2026varsplat}, and SplaTAM~\cite{keetha2024splatam} with method-specific settings. Gaussian-SLAM and VarSplat initialize from two ground-truth poses, while SplaTAM and QuantileGS use no ground-truth motion.
QuantileGS uses up to 40 tracking iterations, 120 initialization iterations, and 60 mapping iterations. Mapping runs every five frames with an eight-keyframe window and a one-million-Gaussian limit. Each successful keyframe uses four UQ fitting epochs with projected Adam and a 5\% RGB-pixel holdout. We retain the background term and gap normalization, targeting 90\% coverage with an online feedback step size of 0.01.
We report equal-scene mean ATE and adjacent-frame translational RPE as RMSE, retaining all frames and using SE(3) alignment without scale correction. FPS includes initialization, tracking, mapping, UQ, and online feedback, excluding input decoding and evaluation. 

\begin{table}[!tb]
    \centering
    \caption{Replica tracking results averaged over eight scenes,
    with seed 0. ATE and RPE are in cm.
    The \colorbox{bestcolor}{best} and
    \colorbox{secondcolor}{second-best} results are highlighted,
    with the best also in bold.}
    \label{tab:slam_results}
    \small
    \setlength{\tabcolsep}{4pt}
    \renewcommand{\arraystretch}{1.12}
    \begin{tabular*}{\columnwidth}{@{\extracolsep{\fill}}lccc@{}}
        \toprule
        Method & ATE $\downarrow$ & RPE $\downarrow$
        & FPS $\uparrow$ \\
        \midrule
        Gaussian-SLAM
        & 0.329 & 0.0878 & \qgsSecond{0.423} \\
        VarSplat
        & \best{\textbf{0.237}} & \qgsSecond{0.0517} & 0.256 \\
        SplaTAM
        & 0.393 & 0.0689 & 0.135 \\
        QuantileGS (Ours)
        & \qgsSecond{0.290} & \best{\textbf{0.0046}}
        & \best{\textbf{1.058}} \\
        \bottomrule
    \end{tabular*}
\end{table}

\noindent\textbf{Results.}
Table~\ref{tab:slam_results} shows that QuantileGS achieves the lowest translational RPE and highest throughput, with the second-lowest ATE after VarSplat.
Online feedback improves mean pre-update RGB holdout coverage from 87.87\% to 89.59\% across 399 evaluated keyframes per sequence, approaching the 90\% target with nearly unchanged throughput. These results assess camera tracking and empirical RGB coverage, rather than global reconstruction quality or trajectory confidence.
Relative to VarSplat, QuantileGS reduces mean translational
RPE by approximately $91.1\%$, while its ATE is $0.053$\, cm
higher. This distinction indicates stronger adjacent-frame translational consistency without the lowest globally aligned trajectory error. More experiments are provided in Appendix~\ref{sec:supp_slam_results}.

%% file: sec/6_conclusion.tex
\section{Conclusion}
\label{sec:conclusion}

We presented QuantileGS, a post-hoc UQ framework that learns ordered error quantiles for a frozen 3DGS renderer. Scale-normalized conformal calibration yields spatially
adaptive RGB-MAE upper bounds with finite-sample marginal coverage under the stated assumptions. Experiments on LF, LLFF, and Mip-NeRF~360 show improved error
ranking over the evaluated baselines. Calibration reduces empirical undercoverage on LLFF and Mip-NeRF~360, while LF bounds remain conservative. Applications to active view selection, Gaussian pruning, and RGB-D SLAM empirically demonstrate the downstream
utility of the calibrated bounds.

%% file: sec/llm_usage_statement.tex
\section*{LLM Usage Statement}

We used ChatGPT for language polishing and grammatical correction. All technical content and conclusions were developed and verified solely by the authors.

%% file: sec/acknowledgment.tex
\section*{Acknowledgment}
The work was supported by the National Key Research and Development Program of China under Grant 2022YFA1003800, the National Natural Science Foundation of China under Grant 62571273, Grant 62125102, and Grant U24B20177, and the Natural Science Foundation of Tianjin under Grant 25JCLMJC01090.

%% file: sec/X_suppl.tex
\clearpage
\setcounter{page}{1}
\renewcommand{\thepage}{S\arabic{page}}
\maketitlesupplementary

\appendix
\numberwithin{equation}{section}

\section{Proof of Theorem~\ref{thm:quantilegs_coverage}}
\label{sec:proof_quantilegs_coverage}

\subsection{Joint Local-Scale Assumptions}

Condition on $\mathcal F_{\mathrm{fit}}$, comprising all
information from $\mathcal V_{\mathrm{fit}}$ used to construct the frozen RGB predictor, quantile fields, and $\beta$, including independent optimization randomness.
The level $\tau$ and floor $\varepsilon>0$ are fixed before
using $\mathcal V_{\mathrm{cal}}$.
Index its $n$ calibration pixels by $i=1,\ldots,n$ and
one new test observation by $i=n+1$.
For functions $\mu,\sigma$ fixed under this conditioning,
assume
\begin{equation}
\begin{aligned}
    E_i&=\mu(X_i)+\sigma(X_i)Z_i, \sigma(X_i)>0,\\
    Z_i&\overset{\mathrm{i.i.d.}}{\sim}F_Z,
    i=1,\ldots,n+1.
\end{aligned}
\label{eq:joint_local_scale_model}
\end{equation}

The entire residual vector $(Z_1,\ldots,Z_{n+1})$ is
independent of the ray vector $(X_1,\ldots,X_{n+1})$.
The fitted fields share the same location and scale:
\begin{equation}
    \hat q_\nu(\boldsymbol x)
    =\mu(\boldsymbol x)+\sigma(\boldsymbol x)\hat z_\nu,
    \nu\in\{\eta,\tau\},
\label{eq:local_scale_fitted_fields}
\end{equation}
where $\hat z_\eta<\hat z_\tau$ are finite constants
determined by fitting. Finally, almost surely,
\begin{equation}
    \Delta(X_i)=\sigma(X_i)(\hat z_\tau-\hat z_\eta)
    \geq\varepsilon, i=1,\ldots,n+1.
\label{eq:inactive_scale_floor}
\end{equation}

This inactive-floor condition is required for the scale
cancellation used below. The three-way view split does
not itself establish these distributional assumptions.

\subsection{Scale Cancellation and Rank Argument}

\noindent\emph{Proof.}
All distributions below are conditional on $\mathcal F_{\mathrm{fit}}$.
After establishing score exchangeability, we apply the standard split-conformal rank argument~\cite{lei2018distributionfree,romano2019conformalized}.
For calibration observations $i=1,\ldots,n$ and the test
observation $i=n+1$, define the random normalized scores
\begin{equation}
    S_i=\frac{E_i-\hat q_\tau(X_i)}{D_\varepsilon(X_i)},
    i=1,\ldots,n+1.
\label{eq:exchangeable_calibration_scores}
\end{equation}

Condition~\eqref{eq:inactive_scale_floor} gives
$D_\varepsilon(X_i)=\Delta(X_i)$ almost surely.
Substituting Eqs.~\eqref{eq:joint_local_scale_model}
and~\eqref{eq:local_scale_fitted_fields} therefore yields
\begin{equation}
\begin{aligned}
    S_i
    &=\frac{\sigma(X_i)(Z_i-\hat z_\tau)}
            {\sigma(X_i)(\hat z_\tau-\hat z_\eta)}\\
    &=\frac{Z_i-\hat z_\tau}{\hat z_\tau-\hat z_\eta}
      =:h(Z_i).
\end{aligned}
\label{eq:normalized_score_local_model}
\end{equation}

The map $h$ is common to all observations and is strictly
increasing because $\hat z_\tau-\hat z_\eta>0$.
It removes all ray-dependent location and scale terms.
For arbitrary real thresholds $t_1,\ldots,t_{n+1}$,
conditional independence of the standardized residuals gives
\begin{equation}
\begin{aligned}
    &\mathbb P\!\left\{
       S_i\leq t_i,\ i=1,\ldots,n+1
       \mid\mathcal F_{\mathrm{fit}}\right\}\\
    &\quad=\prod_{i=1}^{n+1}
       F_Z\!\left(\hat z_\tau+
          (\hat z_\tau-\hat z_\eta)t_i\right).
\end{aligned}
\label{eq:score_joint_distribution}
\end{equation}

The residual vector is also independent of the ray vector
conditional on fitting, so the same joint distribution
holds when additionally conditioning on the rays.
Thus the scores are conditionally i.i.d.\ with a common
ray-invariant distribution, and hence jointly exchangeable.

First suppose that $k_\tau=\lceil(n+1)\tau\rceil\leq n$.
Let $S_{(k_\tau)}$ and $Z_{(k_\tau)}$ denote the
$k_\tau$-th order statistics of the $n$ calibration scores
and standardized residuals, respectively.
Since $h$ is strictly increasing, the calibration rule is
\begin{equation}
    \hat c_\tau=S_{(k_\tau)}=h(Z_{(k_\tau)}).
\label{eq:local_scale_calibration_coefficient}
\end{equation}

For any ray $\boldsymbol x$ on which
$\Delta(\boldsymbol x)\geq\varepsilon$, substituting
this coefficient into Eq.~\eqref{eq:final_quantilegs}
gives
\begin{equation}
\begin{aligned}
    \hat u_\tau(\boldsymbol x)
    ={}&\mu(\boldsymbol x)+\sigma(\boldsymbol x)\hat z_\tau\\
       &+\frac{Z_{(k_\tau)}-\hat z_\tau}
                    {\hat z_\tau-\hat z_\eta}
          \sigma(\boldsymbol x)(\hat z_\tau-\hat z_\eta)\\
    ={}&\mu(\boldsymbol x)+\sigma(\boldsymbol x)Z_{(k_\tau)}.
\end{aligned}
\label{eq:local_scale_calibrated_bound}
\end{equation}

In particular, this identity holds at the test ray almost
surely by Eq.~\eqref{eq:inactive_scale_floor}.

Let $R_{n+1}$ be the rank of the test score among all
$n+1$ scores. Conditional on fitting, break ties using
i.i.d.\ $\operatorname{Uniform}(0,1)$ auxiliary variables
that are independent of all observations.
Exchangeability and this symmetric tie-breaking rule imply
\begin{equation}
    \mathbb P\!\left\{R_{n+1}=r
       \mid\mathcal F_{\mathrm{fit}}\right\}
    =\frac{1}{n+1}, r=1,\ldots,n+1.
\label{eq:uniform_score_rank}
\end{equation}

The event $R_{n+1}\leq k_\tau$ implies
$S_{n+1}\leq S_{(k_\tau)}=\hat c_\tau$.
Consequently,
\begin{equation}
\begin{aligned}
    &\mathbb P\!\left\{S_{n+1}\leq\hat c_\tau
          \mid\mathcal F_{\mathrm{fit}}\right\}\\
    &\quad\geq
       \mathbb P\!\left\{R_{n+1}\leq k_\tau
          \mid\mathcal F_{\mathrm{fit}}\right\}\\
    &\quad=\frac{k_\tau}{n+1}\geq\tau.
\end{aligned}
\label{eq:score_rank_bound}
\end{equation}

Ties can enlarge the non-strict threshold event, so no
continuity assumption on $F_Z$ is needed for this lower bound.

Since $D_\varepsilon(X_{n+1})>0$, the score inequality
is equivalent to the upper-error event:
\begin{equation}
\begin{aligned}
    \{S_{n+1}\leq\hat c_\tau\}
    &=\{E_{n+1}\leq
          \hat q_\tau(X_{n+1})
          +\hat c_\tau D_\varepsilon(X_{n+1})\}\\
    &=\{E_{n+1}\leq\hat u_\tau(X_{n+1})\}.
\end{aligned}
\label{eq:score_coverage_equivalence}
\end{equation}

Combining Eqs.~\eqref{eq:score_rank_bound}
and~\eqref{eq:score_coverage_equivalence} proves
Eq.~\eqref{eq:quantilegs_coverage} for $k_\tau\leq n$.

If $k_\tau>n$, Eq.~\eqref{eq:conformal_coefficient}
sets $\hat c_\tau=+\infty$.
Because $D_\varepsilon(X_{n+1})\geq\varepsilon>0$,
the unclipped upper prediction is $+\infty$ and coverage
is one. The clipped RGB-MAE bound is then one, which
also covers the error because $E_{n+1}\in[0,1]$.

This completes the proof.\hfill$\square$

\subsection{When the Scale Floor Is Active}

Under the same shared location-scale model, the actual score is
\begin{equation}
    S_i=\frac{Z_i-\hat z_\tau}
       {\max\{\hat z_\tau-\hat z_\eta,
                       \varepsilon/\sigma(X_i)\}}.
\label{eq:floored_local_scale_score}
\end{equation}

An active floor can reintroduce ray-dependent scale.
The remaining local-scale assumptions therefore need not
give exchangeable scores. If exchangeability of the actual
floored scores holds separately, the rank argument still
proves marginal coverage because $D_\varepsilon>0$.
The guarantee does not assert coverage at every fixed ray
or simultaneous coverage of all pixels in a view.

\section{Downstream Application Details}
\label{sec:supp_downstream_applications}

This section presents the objectives and computations for the three applications introduced in \Cref{sec:downstream_applications}.

\subsection{Active View Selection}
\label{sec:active_view_selection}

\subsubsection{View Selection Criterion}

AVS selects camera viewpoints to improve scene reconstruction under a limited acquisition budget.
Building on the local information approximation of FisherRF~\cite{jiang2024fisherrf}, we prioritize views that supply missing information about parameters affecting regions with large calibrated error bounds.

At each round, we freeze 3DGS, fit QuantileGS on acquired and reserved fitting views, and calibrate on a disjoint subset. The bounds $u_v(p)\in[0,1]$ include the background contribution and gap normalization. Reserved UQ views are excluded from reconstruction training and the candidate pool.

Let $\mathcal V_0,\mathcal V_c$ denote reconstruction and candidate views, and let $\boldsymbol\theta\in\mathbb R^d$ collect Gaussian geometry, opacity, and active SH parameters.
For view $v$ with $N_v$ pixels, set $n_v=3N_v$ and stack
rendered RGB into $\boldsymbol C_v\in\mathbb R^{n_v}$.
Let $J_v=\partial\boldsymbol C_v/\partial\boldsymbol\theta
\in\mathbb R^{n_v\times d}$ be the RGB Jacobian at the
current model, before output truncation to $[0,1]$.
The diagonal matrix $W_v\in\mathbb R^{n_v\times n_v}$ repeats $u_v(p)$ across RGB channels.
It assigns prediction priorities rather than noise variances.
Define
\begin{equation}
\begin{aligned}
    \boldsymbol F_v&=\frac{\operatorname{diag}(J_v^\top J_v)}{n_v},\\
    \boldsymbol\Lambda&=\sum_{s\in\mathcal V_0}\boldsymbol F_s,\\
    \boldsymbol\omega&=\frac{1}{|\mathcal V_c|}
      \sum_{r\in\mathcal V_c}
      \frac{\operatorname{diag}(J_r^\top W_rJ_r)}{n_r}.
\end{aligned}
    \label{eq:avs_information}
\end{equation}

For parameter $k=1,\ldots,d$, $F_{v,k}$ measures candidate
information, $\Lambda_k$ summarizes acquired information, and
$\omega_k$ measures influence on high-bound regions.

Using a normalized least-squares surrogate with curvature
$J_v^\top J_v/n_v$, a diagonal local Gaussian approximation
gives the parameter-covariance proxy
$\Sigma(\boldsymbol\Lambda)=\operatorname{diag}((\Lambda_k+\lambda)^{-1})$,
with regularization $\lambda>0$.
A first-order rendering approximation propagates it to
$J_r\Sigma(\boldsymbol\Lambda)J_r^\top$.
Its error-weighted mean trace is
\begin{equation}
\begin{aligned}
    \mathcal P_u(\boldsymbol\Lambda)
    &=\frac{1}{|\mathcal V_c|}
      \sum_{r\in\mathcal V_c}
      \frac{\operatorname{tr}
      (W_rJ_r\Sigma(\boldsymbol\Lambda)J_r^\top)}{n_r}\\
    &=\sum_{k=1}^d\frac{\omega_k}{\Lambda_k+\lambda}.
\end{aligned}
    \label{eq:avs_variance_proxy}
\end{equation}

We replace FisherRF's parameter-information objective with
the reduction of this error-weighted prediction-variance
proxy. Within a frozen round, candidate $v$ contributes $\boldsymbol F_v$
and receives the score
\begin{equation}
\begin{aligned}
    \mathcal A(v)&=\mathcal P_u(\boldsymbol\Lambda)-\mathcal P_u(\boldsymbol\Lambda+\boldsymbol F_v)\\
    &=\sum_{k=1}^d
      \frac{\omega_k F_{v,k}}
      {(\Lambda_k+\lambda)(\Lambda_k+\lambda+F_{v,k})}.
\end{aligned}
    \label{eq:avs_acquisition}
\end{equation}

We acquire the highest-scoring view and recompute all
quantities after RGB retraining.
Reconstruction gains are evaluated empirically and do not
follow from the rendering-error coverage guarantee.

\subsubsection{Implementation Details}
We estimate the unweighted and weighted diagonals using
32 Rademacher vector--Jacobian probes per view, applying
$W_v^{1/2}$ to the weighted probes and holding the bounds
fixed during differentiation.
We set $\lambda$ to $0.01$ times the mean positive entry of
$\boldsymbol\Lambda$.
Scoring requires no candidate RGB observations.

After each acquisition, we retrain 3DGS from the same
initialization, progressing from four initial views to
20 reconstruction views.
Each round refits QuantileGS using projected Adam for
200 epochs at learning rate $0.01$.
RGB training uses random backgrounds, while UQ fitting,
calibration, scoring, and evaluation use fixed black
backgrounds.
Reserved UQ views are additional observations.

\subsection{Gaussian Pruning}
\label{sec:gaussian_pruning}

\subsubsection{Construction of Pruning Score}
Gaussian pruning aims to reduce the number of Gaussians while maintaining rendering quality. 
Let $\|\cdot\|$ be the calibrated error norm and $\|\cdot\|_{\mathrm P}$ the norm whose square defines the pruning loss. 
Fix $\kappa>0$ such that $\|\boldsymbol z\|_{\mathrm P}\leq\kappa\|\boldsymbol z\|$.
This separates the calibrated error scale from the pruning criterion through a single conversion factor.

Pruning and refinement change the RGB predictor.
Let $\hat{\boldsymbol y}_0$ and $\hat{\boldsymbol y}_t$ be its
original and current predictions at the same ray, and let
$u^{(0)}$ be the original calibrated bound. We update it as
\begin{equation}
    u^{(t)}
    =\min\left\{e_{\max},\,u^{(0)}
       +\|\hat{\boldsymbol y}_t-\hat{\boldsymbol y}_0\|\right\}.
    \label{eq:bound_transport}
\end{equation}

The triangle inequality and the error limit $e_{\max}$ give
\begin{equation}
\begin{aligned}
    &\|\boldsymbol y-\hat{\boldsymbol y}_0\|\leq u^{(0)} \Longrightarrow\
      \|\boldsymbol y-\hat{\boldsymbol y}_t\|\leq u^{(t)}.
\end{aligned}
    \label{eq:trans43port_event}
\end{equation}

Thus, the update preserves originally covered events at the
same observation, including after data-dependent refinement.

For Gaussian $j$, define its deletion perturbation
$\boldsymbol\delta_j=\hat{\boldsymbol y}_t^{(-j)}
-\hat{\boldsymbol y}_t$ and
$d_j=\|\boldsymbol\delta_j\|_{\mathrm P}$.
Denoting $\boldsymbol r_t=\hat{\boldsymbol y}_t-\boldsymbol y$, on the covered event $\|\boldsymbol r_t\|\leq u^{(t)}$, the triangle inequality bounds the loss increment by
\begin{equation}
\begin{aligned}
    &\|\boldsymbol r_t+\boldsymbol\delta_j\|_{\mathrm P}^2
      -\|\boldsymbol r_t\|_{\mathrm P}^2\\
    &\quad\leq d_j^2
       +2\|\boldsymbol r_t\|_{\mathrm P}d_j\\
    &\quad\leq d_j^2+2\kappa u^{(t)}d_j.
\end{aligned}
    \label{eq:pruning_error_increment}
\end{equation}

For a fixed deletion perturbation $d_j$, a larger error bound $u^{(t)}$ increases the estimated deletion cost.

To define the pruning score, we average the pointwise deletion-cost envelope over reference fitting views $\mathcal V_{\mathrm{ref}}$. 
For any pixel quantity $a(v,p)$, define
\begin{equation}
    \mathbb E_{\mathrm{ref}}[a]
    =\frac{1}{|\mathcal V_{\mathrm{ref}}|}
       \sum_{v\in\mathcal V_{\mathrm{ref}}}
       \frac{1}{N_v}\sum_{p=1}^{N_v}a(v,p),
    \label{eq:pruning_reference_average}
\end{equation}
where $N_v$ is the pixel count of view $v$.
The average envelope contains a squared perturbation term and a residual-dependent term. 
We summarize them using
\begin{equation}
    Q_j=\mathbb E_{\mathrm{ref}}[d_j^2],
    H_j=\mathbb E_{\mathrm{ref}}[(u^{(t)})^2d_j^2].
    \label{eq:pruning_moments}
\end{equation}
where $Q_j$ measures the average squared rendering change, while $H_j$ increases when this change affects pixels with large error bounds. 
Since $\mathbb E_{\mathrm{ref}}[1]=1$, Cauchy-Schwarz applied to $(u^{(t)}d_j)\cdot1$ yields $\mathbb E_{\mathrm{ref}}[u^{(t)}d_j]\leq\sqrt{H_j}$.

Consequently,
\begin{equation}
\begin{aligned}
    &\mathbb E_{\mathrm{ref}}
       [d_j^2+2\kappa u^{(t)}d_j]\\
    &\quad=Q_j+2\kappa\mathbb E_{\mathrm{ref}}[u^{(t)}d_j]\\
    &\quad\leq Q_j+2\kappa\sqrt{H_j}
       =:S_j^{\mathrm{prune}}.
\end{aligned}
    \label{eq:pruning_score}
\end{equation}

We therefore use $S_j^{\mathrm{prune}}$ to rank deletions, prioritizing Gaussians with the smallest scores. 

This favors small rendering changes while accounting for the unknown residual through the calibrated error scale.
In particular, for RGB-MAE calibration and RGB-MSE pruning, $\|\boldsymbol z\|=\|\boldsymbol z\|_1/3$ and $\|\boldsymbol z\|_{\mathrm P}=\|\boldsymbol z\|_2/\sqrt3$, so $\kappa=\sqrt3$ and $S_j^{\mathrm{prune}}=Q_j+2\sqrt{3H_j}$.

\subsubsection{Implementation Details}
Computing $Q_j$ and $H_j$ requires evaluating the deletion perturbation for every Gaussian, which is expensive.
We next estimate these scores without rendering a separate scene for every possible deletion. 
We temporarily multiply each Gaussian's opacity by a gate $g_j$: zero removes the Gaussian, while one keeps its original opacity. 
We evaluate all gates at a common value $\gamma\in(0,1)$. 
Decreasing gate $j$ from $\gamma$ to zero gives the first-order deletion approximation
\begin{equation}
    \boldsymbol\delta_j
    \approx-\gamma\frac{\partial\hat{\boldsymbol y}}{\partial g_j}.
    \label{eq:linearized_deletion}
\end{equation}

The gate derivative therefore supplies a local estimate of how much removing the Gaussian changes the image.

For a reference view with $N$ pixels, we render the RGB image once and independently assign $z_{p,c}=-1$ or $+1$ with equal probability to every pixel and color channel.
Then we form two scalar sums:
\begin{equation}
\begin{aligned}
    L_{\boldsymbol z}
    &=\frac{1}{\sqrt{3N}}\sum_{p,c}z_{p,c}\hat y_{p,c},\\
    L_{\boldsymbol z,u}
    &=\frac{1}{\sqrt{3N}}\sum_{p,c}
        u^{(t)}(p)z_{p,c}\hat y_{p,c}.
\end{aligned}
    \label{eq:pruning_random_sums}
\end{equation}

The first randomly adds or subtracts RGB values; the second also weights each pixel by its error bound.
Keep these bounds fixed and backpropagate each scalar through the renderer. 
Each backward pass returns a derivative for every gate, allowing all Gaussians to be scored together.

To derive the moment estimates, we fix the rendered scene and pixel bounds, and write the RGB derivatives at $g_j=\gamma$ as $a_{p,c,j}=\partial\hat y_{p,c}/\partial g_j$.
Differentiating Eq.~\eqref{eq:pruning_random_sums} gives
\begin{equation}
\begin{aligned}
    \frac{\partial L_{\boldsymbol z}}{\partial g_j}
    &=\frac{1}{\sqrt{3N}}\sum_{p,c}z_{p,c}a_{p,c,j},\\
    \frac{\partial L_{\boldsymbol z,u}}{\partial g_j}
    &=\frac{1}{\sqrt{3N}}\sum_{p,c}
       u^{(t)}(p)z_{p,c}a_{p,c,j}.
\end{aligned}
    \label{eq:pruning_probe_gradients}
\end{equation}

Only the signs are random in this calculation.
Each sign has mean zero and variance one, so independence gives $\mathbb E_{\boldsymbol z}[z_{p,c}z_{p',c'}]=0$ for distinct pixel-channel pairs. 
Expanding the first squared gradient therefore yields
\begin{equation}
\begin{aligned}
    &\mathbb E_{\boldsymbol z}
       \left[\left(\frac{\partial L_{\boldsymbol z}}{\partial g_j}\right)^2\right]\\
    &\quad=\frac{1}{3N}\sum_{p,c}\sum_{p',c'}
       a_{p,c,j}a_{p',c',j}
       \mathbb E_{\boldsymbol z}[z_{p,c}z_{p',c'}]\\
    &\quad=\frac{1}{3N}\sum_{p,c}a_{p,c,j}^2.
\end{aligned}
    \label{eq:pruning_probe_second_moment}
\end{equation}

All cross-terms vanish, leaving the sum of squared RGB derivatives. 
The same argument with the fixed pixel weights gives
\begin{equation}
\begin{aligned}
    &\mathbb E_{\boldsymbol z}
       \left[\left(\frac{\partial L_{\boldsymbol z,u}}
                         {\partial g_j}\right)^2\right]\\
    &\qquad=\frac{1}{3N}\sum_{p,c}
       (u^{(t)}(p))^2a_{p,c,j}^2.
\end{aligned}
    \label{eq:pruning_weighted_probe_second_moment}
\end{equation}

By Eq.~\eqref{eq:linearized_deletion}, multiplying these expectations by $\gamma^2$ gives the unweighted and bound-weighted moments of the linearized deletion response.
The factor $1/(3N)$ already averages over RGB channels and pixels.
We approximate each expectation by averaging $K$ independent random draws, giving the per-view estimates
\begin{equation}
\begin{aligned}
    \widehat Q_{j,v}
    &=\frac{\gamma^2}{K}\sum_{k=1}^{K}
        \left(\frac{\partial L_{\boldsymbol z_k}}
                   {\partial g_j}\right)^2,\\
    \widehat H_{j,v}
    &=\frac{\gamma^2}{K}\sum_{k=1}^{K}
        \left(\frac{\partial L_{\boldsymbol z_k,u}}
                   {\partial g_j}\right)^2.
\end{aligned}
    \label{eq:randomized_deletion_moment}
\end{equation}

Finally, average these estimates equally over reference views and substitute them into Eq.~\eqref{eq:pruning_score} to rank
Gaussians.
Algorithm~\ref{alg:pruning_computation} summarizes the computation.

We alternate deletion of low-scoring Gaussians with refinement and rescoring as visibility changes.
The scores use approximate deletion responses, whereas the bound update in Eq.~\eqref{eq:bound_transport} uses the actual RGB predictions before and after pruning.

\begin{algorithm}[t]
    \caption{Gaussian Pruning Score Estimation}
    \label{alg:pruning_computation}
    \small
    \algrenewcommand{\algorithmicrequire}{\textbf{Input:}}
    \algrenewcommand{\algorithmicensure}{\textbf{Output:}}
    \begin{algorithmic}[1]
        \Require Frozen Gaussian scene $\mathcal G$; reference views
        $\mathcal V_{\mathrm{ref}}$; fixed RGB-MAE bounds
        $\{u_v^{(t)}\}$; gate value $\gamma\in(0,1)$;
        number of probes $K\geq1$.
        \Ensure Pruning scores $\{S_j^{\mathrm{prune}}\}$.
        \State Set $g_j\gets\gamma$ for every Gaussian $j$.
        \State Initialize $\widehat Q_j\gets0$, $\widehat H_j\gets0$ for all $j$.
        \For{each view $v\in\mathcal V_{\mathrm{ref}}$}
            \State Render its RGB image once using the opacity gates.
            \For{$k=1,\ldots,K$}
                \State Draw independent signs
                $z_{p,c}\in\{-1,+1\}$ uniformly.
                \State Form $L_{\boldsymbol z}$ and $L_{\boldsymbol z,u}$
                using Eq.~\eqref{eq:pruning_random_sums}.
                \State Backpropagate each scalar to all gates.
                \State For all $j$, accumulate
                $\widehat Q_j\gets\widehat Q_j+
                \gamma^2(\partial L_{\boldsymbol z}/\partial g_j)^2$.
                \State For all $j$, accumulate
                $\widehat H_j\gets\widehat H_j+
                \gamma^2(\partial L_{\boldsymbol z,u}/\partial g_j)^2$.
            \EndFor
        \EndFor
        \For{each Gaussian $j$}
            \State $\widehat Q_j\gets
                \widehat Q_j/(K|\mathcal V_{\mathrm{ref}}|)$.
            \State $\widehat H_j\gets
                \widehat H_j/(K|\mathcal V_{\mathrm{ref}}|)$.
            \State $S_j^{\mathrm{prune}}\gets
                \widehat Q_j+2\sqrt{3\widehat H_j}$.
        \EndFor
        \State \Return $\{S_j^{\mathrm{prune}}\}$; smaller scores are pruned first.
    \end{algorithmic}
\end{algorithm}

\subsection{Simultaneous Localization and Mapping}
\label{sec:slam_method}

\subsubsection{RGB-D Tracking and Mapping}

SLAM estimates a camera trajectory while reconstructing the scene observed along that trajectory. 
Given RGB images $\boldsymbol y_t$, depth maps $D_t$, and known camera intrinsics $\boldsymbol K$, which specify camera projection, we estimate camera poses $T_t\in\mathrm{SE}(3)$ and incrementally build a Gaussian map $\mathcal G$.
Each pose transforms 3D points from camera to world coordinates. 
Tracking aligns each incoming frame with the previously observed scene to estimate its pose; mapping then uses this pose to update scene geometry and appearance.

We adopt the alternating tracking-mapping organization of Gaussian-SLAM~\cite{yugay2023gaussianslam}.
Gaussian-SLAM estimates poses against a rendered Gaussian submap, while VarSplat~\cite{tran2026varsplat} additionally uses learned appearance variance to weight photometric residuals. 
Our integration uses QuantileGS to construct reference images with calibrated rendering-error bounds.
New frames are aligned with these cached RGB-D references by reprojection, and their bounds determine the robust scales of the photometric tracking loss.
Thus, Gaussian rendering supplies the reference appearance, while pose optimization operates on image correspondences.

We reserve an RGB pixel subset in each frame for calibration; only the remaining \emph{fitting pixels} enter RGB tracking, map fitting, quantile learning, and reference correction.
Depth remains available for geometric alignment.
We set $T_0=I$ and initialize $\mathcal G$ from the first frame's fitting RGB-D. 
Selected frames (\emph{keyframes}) supply reference observations for subsequent tracking and map refinement. 
Let $\mathcal W$ denote a recent-keyframe window used for fitting, and let $\mathcal R$ index the recent reference keyframes used for tracking.
The input frame-index sets $\mathcal K$ and $\mathcal M$ specify keyframes and map-refinement frames, respectively, with $0\in\mathcal K$.

\subsubsection{Calibrated Reference Construction}

After updating the map at keyframe $k$, we freeze a copy, fit the ordered quantile fields on fitting RGB errors from $\mathcal W$, and calibrate them on the fresh reserved RGB
of frame $k$. 
Rendering this frozen predictor at $T_k$ gives colors $\hat{\boldsymbol y}_{k,p}$ and bounds $u_{k,p}$ on $\|\boldsymbol y_{k,p}-\hat{\boldsymbol y}_{k,p}\|$.
We then incorporate the observed fitting colors into the saved reference, using the bound as a correction budget.
This allows the reference to use measured appearance while controlling its departure from the calibrated rendering.

Similar with \Cref{sec:gaussian_pruning}, we choose $\kappa_{\mathrm{ch}}$ such that $|a_c|\leq\kappa_{\mathrm{ch}}\|\boldsymbol a\|$ for every RGB error vector $\boldsymbol a$ and channel $c$.
A covered RGB error therefore implies a channel error at most $b_{k,p}=\kappa_{\mathrm{ch}}u_{k,p}$.
At fitting pixels, the corrected reference is
\begin{equation}
\begin{aligned}
    \widetilde y_{k,p,c}
    ={}&\hat y_{k,p,c}+\Pi_{[-b_{k,p},\,b_{k,p}]}
        (y_{k,p,c}-\hat y_{k,p,c}),
\end{aligned}
    \label{eq:slam_reference_correction}
\end{equation}
where $\Pi_I(v)=\arg\min_{s\in I}(s-v)^2$ denotes projection onto interval $I$. 
Intuitively, each channel moves toward its observation by at most $b_{k,p}$; on the covered event it reaches that observation. 
Reserved pixels retain their rendered colors, and the Gaussian parameters remain fixed.
As in Eq.~\eqref{eq:bound_transport}, we attach the conservatively transported bound
\begin{equation}
    \widetilde u_{k,p}
    =\min\left\{e_{\max},\,u_{k,p}
       +\|\widetilde{\boldsymbol y}_{k,p}
                -\hat{\boldsymbol y}_{k,p}\|\right\}.
    \label{eq:slam_reference_transport}
\end{equation}

The triangle inequality preserves every originally covered observation under this change of reference.
We cache $(\widetilde{\boldsymbol y}_k,\widetilde u_k,D_k,T_k)$ and the fitting mask; subsequent map updates leave this
reference fixed.
Our implementation uses RGB-MAE, for which $\kappa_{\mathrm{ch}}=3$ and $e_{\max}=1$.

\subsubsection{Uncertainty-Aware Tracking and Map Updates}

For an incoming frame $t$, we estimate its pose by aligning
its observed RGB-D with the references in $\mathcal R$.
Depth converts pixel $p$ into a 3D point, and a candidate
pose $T$ projects that point into reference image $k$:
\begin{equation}
\begin{aligned}
    \boldsymbol X_t(p)&=D_t(p)\boldsymbol K^{-1}\bar{\boldsymbol p},\\
    w_{k\leftarrow t}(p;T)
      &=\pi_{\boldsymbol K}\bigl(T_k^{-1}T\boldsymbol X_t(p)\bigr),
\end{aligned}
    \label{eq:slam_pixel_correspondence}
\end{equation}
where $p_x,p_y$ are the horizontal and vertical pixel
coordinates, $\bar{\boldsymbol p}=(p_x,p_y,1)^\top$, and
$\pi_{\boldsymbol K}$ denotes camera projection.
Where $\boldsymbol X_t(p)$ is expressed in the current
camera's coordinates; $T_k^{-1}T$ transforms it into
reference-camera coordinates.
The resulting pixel location $w_{k\leftarrow t}(p;T)$
determines which reference color to compare with the
current observation:
\begin{equation}
    r_{k,p,c}(T)
    =\widetilde y_{k,c}\bigl(w_{k\leftarrow t}(p;T)\bigr)
      -y_{t,p,c}.
    \label{eq:slam_photometric_residual}
\end{equation}

The error bound at the matched reference pixel determines
how strongly this color residual influences tracking.
We set its robust scale $\ell_{k,p}$ from
$\kappa_{\mathrm{ch}}\widetilde u_k
(w_{k\leftarrow t}(p;T))$, with a positive floor to avoid
zero scales. The photometric objective is
\begin{equation}
    \mathcal L_{\mathrm{rgb}}(T)
    =\sum_{k\in\mathcal R}\sum_{p\in\Omega_k}
      a_{k,p}\sum_{c=1}^3
      \rho_{\mathrm H}\!\left(
        \frac{r_{k,p,c}(T)}{\ell_{k,p}}\right),
    \label{eq:slam_photometric_objective}
\end{equation}
where $\Omega_k$ contains visible, depth-consistent fitting
matches, and $a_{k,p}\geq0$ combines fitting-pixel weights
with per-reference normalization.

The Huber penalty $\rho_{\mathrm H}(z)$ equals $z^2/2$
for $|z|\leq1$ and $|z|-1/2$ otherwise.
Thus, the calibrated bound supplies a local error scale,
while the linear tail limits the influence of large
photometric mismatches.
Reserved RGB is excluded before image resizing or sampling.

Depth provides a complementary geometric constraint.
We use $\mathcal L_{\mathrm{depth}}(T)$, a robust
point-to-plane alignment cost: it penalizes the distance
between transformed current depth points and the
corresponding reference surfaces.
This term uses observed depth and a separate scale in
meters; RGB error bounds are used only in the photometric
term. We estimate $T_t$ by minimizing
\begin{equation}
    \mathcal L_{\mathrm{track}}(T)
    =\mathcal L_{\mathrm{rgb}}(T)
       +\lambda_d\mathcal L_{\mathrm{depth}}(T),
    \lambda_d>0.
    \label{eq:slam_pose_objective}
\end{equation}

We initialize from recent camera motion and apply
Gauss-Newton updates from coarse to fine.
We refresh matches and sampled scales between iterations,
keeping masks, scales, and weights fixed when testing
an update.

After tracking, accepted frames in \(\mathcal{M}\cup\mathcal{K}\) update \(\mathcal{G}\) with the estimated poses fixed. Map refinement minimizes fitting color and depth errors over the current frame and \(\mathcal{W}\). At accepted keyframes, newly observed fitting RGB-D initializes additional Gaussians before refinement, and a calibrated reference is constructed afterward for future tracking, as shown in Algorithm~\ref{alg:slam}. The calibration statement concerns rendering errors, and we evaluate its effect on trajectory accuracy empirically.

In Algorithm~\ref{alg:slam}, $\mathcal F_t$ and $\mathcal H_t$ denote fitting and reserved RGB pixels.
$\mathsf{MapUpdate}$ performs the map update described above, and $\mathsf{Recent}$ maintains the corresponding recent window. For a score set $S$ of size $n$, $\mathsf Q^+_{1-a}(S)$ denotes the $\lceil(n+1)(1-a)\rceil$-th order statistic of $S\cup\{+\infty\}$. Within reference construction, pixel-wise quantities refer to the current frame.

\begin{algorithm}[t]
    \caption{QuantileGS RGB-D SLAM with Online Calibration}
    \label{alg:slam}
    \small
    \makeatletter
    \def\theHALG@line{\thealgorithm.\arabic{ALG@line}}
    \makeatother
    \algrenewcommand{\algorithmicrequire}{\textbf{Input:}}
    \algrenewcommand{\algorithmicensure}{\textbf{Output:}}
    \begin{algorithmic}[1]
        \Require RGB-D frames
        $\{(\boldsymbol y_t,D_t,\mathcal F_t,\mathcal H_t)\}$,
        intrinsics $\boldsymbol K$;
        quantile levels $(\eta,\tau)$ and floor $\epsilon$;
        nominal miscoverage $\alpha$;
        feedback parameters $\gamma,[a_{\min},a_{\max}]$;
        frame sets $\mathcal K,\mathcal M$.
        \Ensure Camera poses $\{T_t\}$ and Gaussian map $\mathcal G$.

        \State $\mathcal G,\mathcal W,\mathcal R\gets\varnothing$;
        $T_0\gets I$; $a\gets\alpha$.

        \For{each frame $t$}
            \State For $t>0$, estimate $T_t$ against $\mathcal R$
            by minimizing
            $\mathcal L_{\mathrm{rgb}}(T)
            +\lambda_d\mathcal L_{\mathrm{depth}}(T)$.
            \State If tracking is rejected,
            $T_t\gets T_{t-1}$.

            \If{$t\in\mathcal K$ and a predictor is available}
                \State $\displaystyle
                \widehat C_t\gets
                \frac{1}{|\mathcal H_t|}
                \sum_{p\in\mathcal H_t}
                \mathbf 1\!\left\{
                    \|\hat{\boldsymbol y}_{t,p}^{-}
                    -\boldsymbol y_{t,p}\|
                    \leq u_{t,p}^{-}
                \right\}$.
                \State $\displaystyle
                a\gets\Pi_{[a_{\min},a_{\max}]}
                \bigl(a+\gamma[\widehat C_t-(1-\alpha)]\bigr)$.
            \EndIf

            \If{$t=0$ or tracking is accepted}
                \State If $t\in\mathcal M\cup\mathcal K$,
                $\mathcal G\gets
                \mathsf{MapUpdate}(\mathcal G,\mathcal W,t)$.

                \If{$t\in\mathcal K$}
                    \State $\mathcal W\gets
                    \mathsf{Recent}(\mathcal W,t)$;
                    $\mathcal G^\star\gets
                    \mathsf{FreezeCopy}(\mathcal G)$.
                    \State $\phi\gets
                    \mathsf{FitQuantiles}
                    (\mathcal G^\star,\mathcal W;\eta,\tau)$
                    using fitting RGB only.
                    \State Render
                    $(\hat{\boldsymbol y},\hat q_\eta,\hat q_\tau)$
                    from $(\mathcal G^\star,\phi)$ at $T_t$.

                    \State $\displaystyle
                    d_p\gets
                    \max\{\hat q_{\tau,p}-\hat q_{\eta,p},
                    \epsilon\}$.
                    \State $\displaystyle
                    s_p\gets
                    \frac{
                        \|\boldsymbol y_{t,p}
                        -\hat{\boldsymbol y}_p\|
                        -\hat q_{\tau,p}
                    }{d_p},
                    \quad p\in\mathcal H_t$.
                    \State $\displaystyle
                    \hat c\gets
                    \mathsf Q^+_{1-a}
                    (\{s_p:p\in\mathcal H_t\})$.
                    \State $\displaystyle
                    u_p\gets
                    \operatorname{clip}_{[0,e_{\max}]}
                    (\hat q_{\tau,p}+\hat c\,d_p)$.

                    \State $\widetilde{\boldsymbol y}
                    \gets\hat{\boldsymbol y}$;
                    $b_p\gets\kappa_{\mathrm{ch}}u_p$.
                    \State For $p\in\mathcal F_t$ and each channel $c$:
                    \Statex \hspace{\algorithmicindent}
                    $\displaystyle
                    \widetilde y_{p,c}\gets
                    \hat y_{p,c}
                    +\Pi_{[-b_p,b_p]}
                    (y_{t,p,c}-\hat y_{p,c})$.
                    \State $\displaystyle
                    \widetilde u_p\gets
                    \min\!\left\{
                        e_{\max},\,
                        u_p+
                        \|\widetilde{\boldsymbol y}_p
                        -\hat{\boldsymbol y}_p\|
                    \right\}$.
                    \State $\mathcal R\gets
                    \mathsf{Recent}\!\left(
                        \mathcal R,
                        (\widetilde{\boldsymbol y},
                        \widetilde u,D_t,T_t,\mathcal F_t)
                    \right)$.
                \EndIf
            \EndIf
        \EndFor

        \State \Return $\{T_t\},\mathcal G$.
    \end{algorithmic}
\end{algorithm}

\subsubsection{Online Calibration}
\label{sec:slam_online_calibration}

We adapt the calibration level using observed coverage as
the camera moves, and the map evolves. The nominal miscoverage level remains fixed at $\alpha=0.1$. We maintain an effective calibration level $a_t$, initialized
with $a_0=\alpha$.

After tracking an evaluated keyframe $t$, but before updating the map or quantile fields, we measure coverage on its reserved RGB pixels $\mathcal H_t$:
\begin{equation}
    \widehat C_t
    =
    \frac{1}{|\mathcal H_t|}
    \sum_{p\in\mathcal H_t}
    \mathbf{1}\!\left\{
        \|\hat{\boldsymbol y}_{t,p}^{-}
          -\boldsymbol y_{t,p}\|
        \leq u_{t,p}^{-}
    \right\}.
    \label{eq:slam_preupdate_coverage}
\end{equation}
where $\hat{\boldsymbol y}_{t,p}^{-}$ is rendered from the
map before its update, and $u_{t,p}^{-}$ is supplied by the
existing calibrated predictor, transported to this map
using Eq.~\eqref{eq:bound_transport} when necessary.

We update the effective miscoverage level by
\begin{equation}
    a_{t+1}
    =
    \Pi_{[a_{\min},a_{\max}]}
    \left(
        a_t+\gamma
        \bigl[\widehat C_t-(1-\alpha)\bigr]
    \right),
    \label{eq:slam_online_feedback}
\end{equation}
where $\gamma=0.01$, $a_{\min}=0.001$, and $a_{\max}=0.5$.
On frames without feedback, $a_{t+1}=a_t$. Undercoverage decreases $a_t$, requesting a higher calibration
quantile; overcoverage increases it.

At the next successful quantile-field refresh, the latest
effective level replaces $\alpha$ in the conformal order
statistic. The fitting levels $\eta,\tau$ remain fixed.
The current keyframe's reserved RGB pixels subsequently
calibrate the refreshed predictor, which supplies references for future tracking.
Feedback does not revise the completed tracking estimate or
its recorded pre-update coverage.
This mechanism provides empirical coverage control and does
not establish a per-frame finite-sample coverage guarantee.

\section{Uncertainty Quantification Results}
\label{sec:supp_uq_results}

We provide additional results on five selected scenes not shown
in the main qualitative comparison: LLFF flower and room,
Mip-NeRF~360 bonsai and stump, and LF torch.
QuantileGS-U and QuantileGS-C denote predictions before and
after conformal calibration, respectively.

\subsection{Selected-Scene Quantitative Results}
\label{sec:supp_uq_selected}

\begin{figure*}[t]
    \centering
    \includegraphics[width=1.0\textwidth]{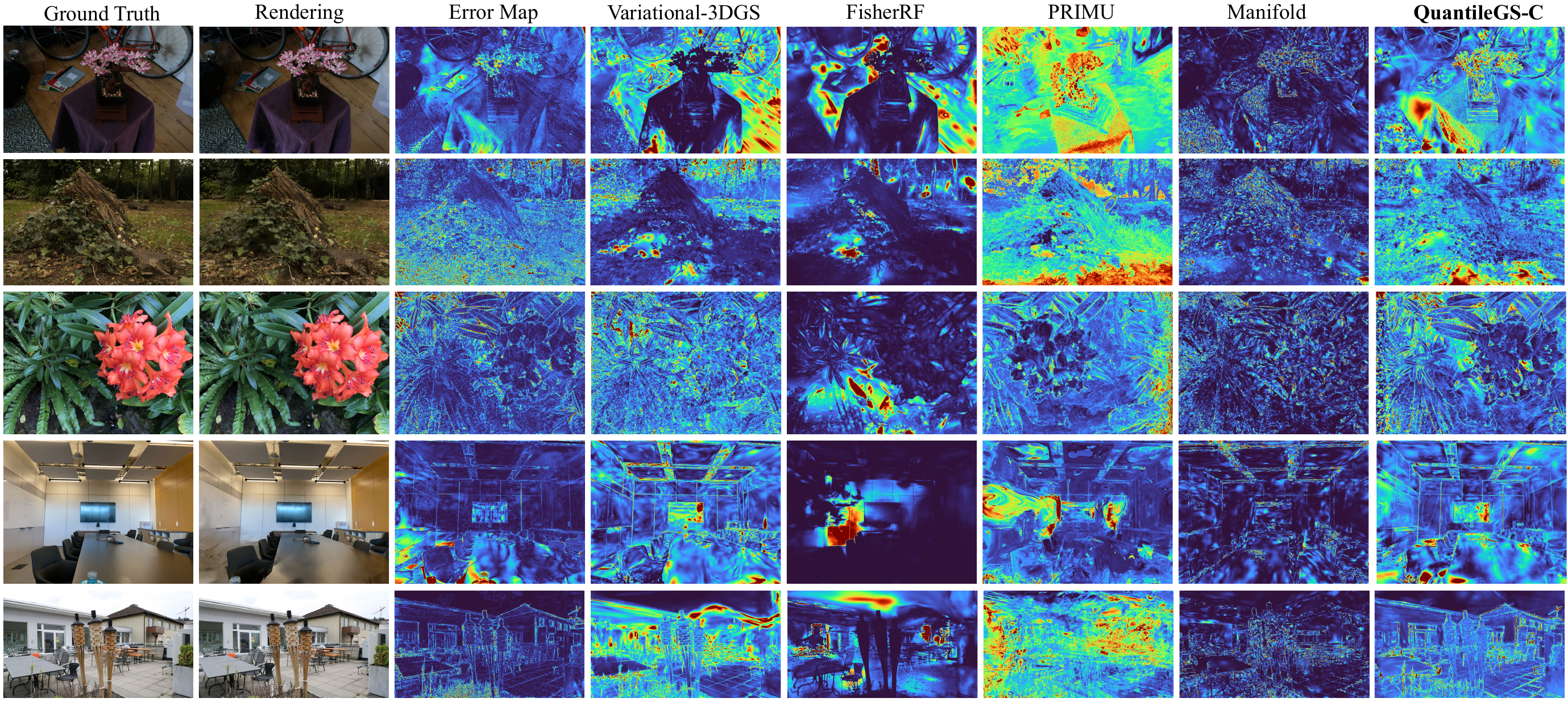}
    \caption{
        Additional qualitative comparisons of
        Variational-3DGS~\cite{li2024variational},
        FisherRF~\cite{jiang2024fisherrf},
        PRIMU~\cite{gottwald2026primu},
        Manifold~\cite{lyu2024manifold}, and QuantileGS-C (Ours).
        Each scene shows uncertainty maps.
        Scenes from top to bottom:  Mip-NeRF~360 bonsai and stump, 
        LLFF flower and room, and LF torch.
    }
    \label{fig:supp_uq_selected}
\end{figure*}

Table~\ref{tab:supp_uq_selected} reports AUSE and Pearson correlation,
retaining each baseline's original
evaluation conventions.
QuantileGS-C outperforms all baselines in both ranking metrics
on the selected LLFF and Mip-NeRF~360 scenes.
On torch, it achieves the lowest AUSE alongside QuantileGS-U, while Variational-3DGS retains higher Pearson correlation. Fig.~\ref{fig:supp_uq_selected} compares RGB renderings, rendering errors, and uncertainty maps on the same five scenes. 

\begin{table*}[t]
    \centering
    \caption{
        UQ results on five additional selected scenes.
        The \colorbox{bestcolor}{best},
        \colorbox{secondcolor}{second-best}, and
        \colorbox{thirdcolor}{third-best} AUSE and Pearson results are highlighted, with the best also in bold.
    }
    \label{tab:supp_uq_selected}
    \small
    \renewcommand{\arraystretch}{1.15}
    \setlength{\tabcolsep}{5pt}
    \resizebox{\textwidth}{!}{
    \begin{tabular}{l*{10}{c}}
        \toprule
        \multirow{3}{*}{Method}
        & \multicolumn{4}{c}{LLFF}
        & \multicolumn{4}{c}{Mip-NeRF~360}
        & \multicolumn{2}{c}{LF} \\
        \cmidrule(lr){2-5}
        \cmidrule(lr){6-9}
        \cmidrule(lr){10-11}

        & \multicolumn{2}{c}{Flower}
        & \multicolumn{2}{c}{Room}
        & \multicolumn{2}{c}{Bonsai}
        & \multicolumn{2}{c}{Stump}
        & \multicolumn{2}{c}{Torch} \\
        \cmidrule(lr){2-3}
        \cmidrule(lr){4-5}
        \cmidrule(lr){6-7}
        \cmidrule(lr){8-9}
        \cmidrule(lr){10-11}

        & AUSE $\downarrow$ & Pears. $\uparrow$
        & AUSE $\downarrow$ & Pears. $\uparrow$
        & AUSE $\downarrow$ & Pears. $\uparrow$
        & AUSE $\downarrow$ & Pears. $\uparrow$
        & AUSE $\downarrow$ & Pears. $\uparrow$ \\
        \midrule

        Variational-3DGS
        & 0.379 & 0.226
        & \third{0.357} & \third{0.279}
        & 0.473 & 0.072
        & 0.526 & 0.123
        & \qgsSecond{0.223} & \best{\textbf{0.434}} \\

        FisherRF
        & 0.328 & -0.094
        & 0.382 & -0.070
        & 0.503 & -0.071
        & \third{0.356} & -0.020
        & 0.700 & -0.051 \\

        PRIMU
        & \third{0.257} & \third{0.406}
        & 0.375 & 0.207
        & 0.364 & 0.264
        & 0.435 & 0.134
        & \third{0.345} & 0.358 \\

        Manifold
        & 0.462 & 0.182
        & 0.468 & 0.179
        & \third{0.350} & \third{0.302}
        & 0.483 & \third{0.144}
        & 0.514 & 0.126 \\

        QuantileGS-U (Ours)
        & \best{\textbf{0.144}} & \best{\textbf{0.458}}
        & \qgsSecond{0.268} & \best{\textbf{0.336}}
        & \best{\textbf{0.299}} & \qgsSecond{0.340}
        & \best{\textbf{0.306}} & \best{\textbf{0.481}}
        & \best{\textbf{0.218}} & \third{0.385} \\

        QuantileGS-C (Ours)
        & \qgsSecond{0.156} & \qgsSecond{0.452}
        & \best{\textbf{0.225}} & \qgsSecond{0.333}
        & \qgsSecond{0.300} & \best{\textbf{0.343}}
        & \qgsSecond{0.309} & \qgsSecond{0.450}
        & \best{\textbf{0.218}} & \qgsSecond{0.386} \\
        \bottomrule
    \end{tabular}
    }
\end{table*}

\subsection{Empirical Coverage and Bound Tightness}
\label{sec:supp_uq_coverage_tightness}

We compare QuantileGS-U and QuantileGS-C on all scenes using the same configurations and test views as in the UQ evaluation. Coverage is the fraction of test pixels whose RGB-MAE does not exceed the predicted upper bound.
Mean bound averages these predictions over the same pixels.
The nominal coverage level is $90\%$, with $\alpha=0.1$.
Both quantities are computed by pooling test pixels within each scene. Dataset averages weight scenes equally.
A smaller mean bound indicates tighter predictions but must be interpreted together with coverage.

\begin{table*}[t]
\centering
\caption{\textbf{Coverage and bound size before and after calibration.}
The nominal coverage is $90\%$. Mean bounds are in RGB-MAE units
for RGB values in $[0,1]$. Shaded rows are equal-scene means.}
\label{tab:supp_uq_coverage_tightness}

\small
\setlength{\tabcolsep}{5pt}
\renewcommand{\arraystretch}{1.08}
\sisetup{
    detect-weight=true,
    detect-family=true,
    group-digits=none
}

\begin{tabularx}{\textwidth}{@{}
    >{\centering\arraybackslash}X
    >{\raggedright\arraybackslash}p{0.12\textwidth}
    *{2}{S[table-format=2.2,table-column-width=0.16\textwidth]}
    *{2}{S[table-format=1.4,table-column-width=0.16\textwidth]}
    @{}}
\toprule
\multirow{2}{*}{Dataset}
& \multirow{2}{*}{Scene}
& \multicolumn{2}{c}{Coverage (\%)}
& \multicolumn{2}{c}{Mean bound} \\
\cmidrule(lr){3-4}
\cmidrule(lr){5-6}
& & {QuantileGS-U} & {\textbf{QuantileGS-C}}
  & {QuantileGS-U} & {\textbf{QuantileGS-C}} \\
\midrule

\multirow{9}{*}{LLFF}
& fern     & 57.29 & 89.57 & 0.0290 & 0.0953 \\
& flower   & 73.49 & 93.48 & 0.0337 & 0.0713 \\
& fortress & 66.36 & 89.53 & 0.0254 & 0.0503 \\
& horns    & 82.40 & 89.39 & 0.0408 & 0.0529 \\
& leaves   & 65.55 & 89.96 & 0.0907 & 0.2319 \\
& orchids  & 51.18 & 88.94 & 0.0457 & 0.1837 \\
& room     & 78.21 & 91.39 & 0.0206 & 0.0373 \\
& trex     & 81.97 & 84.59 & 0.0448 & 0.0490 \\
\rowcolor{qgsSummary}
& \textbf{Average} & 69.56 & 89.61 & 0.0413 & 0.0965 \\

\addlinespace[4pt]

\multirow{10}{*}{Mip-NeRF~360}
& bicycle  & 75.84 & 94.91 & 0.0596 & 0.1345 \\
& bonsai   & 87.15 & 91.40 & 0.0350 & 0.0431 \\
& counter  & 85.30 & 91.59 & 0.0395 & 0.0545 \\
& flowers  & 81.54 & 92.73 & 0.0957 & 0.1485 \\
& garden   & 83.49 & 90.44 & 0.0489 & 0.0642 \\
& kitchen  & 87.30 & 91.16 & 0.0396 & 0.0478 \\
& room     & 83.62 & 90.82 & 0.0335 & 0.0474 \\
& stump    & 64.19 & 91.48 & 0.0444 & 0.1351 \\
& treehill & 72.23 & 94.82 & 0.0656 & 0.1606 \\
\rowcolor{qgsSummary}
& \textbf{Average} & 80.07 & 92.15 & 0.0513 & 0.0929 \\

\addlinespace[4pt]

\multirow{5}{*}{LF}
& africa & 96.66 & 97.12 & 0.0696 & 0.0748 \\
& basket & 94.84 & 93.94 & 0.0816 & 0.0736 \\
& statue & 98.70 & 98.66 & 0.0638 & 0.0625 \\
& torch  & 95.07 & 95.83 & 0.0584 & 0.0656 \\
\rowcolor{qgsSummary}
& \textbf{Average} & 96.32 & 96.39 & 0.0684 & 0.0691 \\

\bottomrule
\end{tabularx}
\end{table*}

\paragraph{Coverage and bound size.}
Table~\ref{tab:supp_uq_coverage_tightness} shows that calibration substantially reduces undercoverage on LLFF and Mip-NeRF~360. Mean coverage increases from 69.56\% to 89.61\% on LLFF and from 80.07\% to 92.15\% on Mip-NeRF~360. These gains accompany wider bounds, whose means increase from 0.0413 to 0.0965 and from 0.0513 to 0.0929, respectively. The correction varies across scenes.
For example, stump coverage rises from 64.19\% to 91.48\%
as its mean bound increases from 0.0444 to 0.1351.
Bonsai requires a smaller adjustment, from 0.0350 to 0.0431, to raise coverage from 87.15\% to 91.40\%.

\paragraph{Scene-level variation.}
All nine Mip-NeRF~360 scenes exceed the nominal coverage level after calibration, with empirical coverage ranging from 90.44\% to 94.91\%. LLFF remains less uniform.
Although its mean coverage approaches 90\%, six scenes remain below this level, with the largest shortfall on trex at 84.59\%.
The dataset mean therefore does not establish adequate coverage for every scene.

\paragraph{Conservative predictions on LF.}
LF already exceeds the nominal level before calibration.
Mean coverage changes only slightly, from 96.32\% to 96.39\%, while the mean bound changes from 0.0684 to 0.0691. Calibration tightens the bounds on basket and statue, whose coverage remains above 90\%, but widens them on africa and torch. Thus, calibration does not uniformly enlarge the bounds or bring every scene closer to the nominal level. These results report empirical coverage under the evaluated protocols rather than conditional coverage for individual views or pixels.

\section{Downstream Application Results}
\label{sec:supp_downstream_results}

\setlength{\floatsep}{8pt plus 2pt minus 2pt}
\setlength{\textfloatsep}{8pt plus 2pt minus 2pt}
\setlength{\dblfloatsep}{8pt plus 2pt minus 2pt}
\setlength{\dbltextfloatsep}{8pt plus 2pt minus 2pt}
\setlength{\abovecaptionskip}{4pt}
\setlength{\belowcaptionskip}{0pt}

We provide per-scene results and qualitative comparisons for Gaussian pruning and SLAM. All experiments use seed~0, and dataset averages weight scenes equally. Performance values are ranked before rounding. The \colorbox{bestcolor}{best}, \colorbox{secondcolor}{second-best}, and
\colorbox{thirdcolor}{third-best} results are highlighted,
with the best also in bold. Coverage, bound widths, and model sizes are reported as diagnostic quantities without ranking.

\subsection{Gaussian Pruning}
\label{sec:supp_pruning_results}

\paragraph{Per-scene comparison.}
We evaluate all nine Mip-NeRF~360 scenes at 20\% and 10\% retention.
All methods use the original checkpoints and test views, and rendering metrics are computed from floating-point RGB images. QuantileGS attains the highest mean PSNR at both retention levels: it ranks first on five scenes at 20\% retention and six at 10\%, and its mean PSNR exceeds that of PUP by 0.053 and 0.269\,dB, respectively.

\paragraph{Model size and transported bounds.}
Table~\ref{tab:supp_pruning_counts} reports Gaussian counts, QuantileGS coverage, and mean RGB-MAE bounds. Counts are identical across methods at each retention level. The mean count decreases from approximately 2.44 million to 488 thousand and 244 thousand.
Coverage increases from 92.14\% to
96.44\% and 96.72\%, while mean bounds widen from 0.0929 to
0.1125 and 0.1159. No originally covered test pixel becomes
uncovered under the transported bound in the saved evaluations.
The higher coverage accompanies more conservative bounds and does not establish improved calibration or a PSNR guarantee. Evaluating transported bounds additionally requires the original renderer and its quantile fields.

\begin{table*}[t]
\centering
\caption{\textbf{Compression, coverage, and bound size after pruning.}
The $20\%$ and $10\%$ columns denote Gaussian retention levels.
Counts are identical across the three compared methods.
QuantileGS bounds for pruned models are transported from the original
calibrated model. Coverage (\%) and mean RGB-MAE bounds pool test pixels
within each scene. Average weights scenes equally.
All evaluated coverage-implication violation counts are zero.}
\label{tab:supp_pruning_counts}
\label{tab:supp_pruning_coverage}
\small
\setlength{\tabcolsep}{3pt}
\renewcommand{\arraystretch}{1.13}
\sisetup{detect-weight=true,detect-family=true,group-digits=none}
\begin{tabularx}{\textwidth}{X
    S[table-format=7.1,table-column-width=0.126\textwidth,
      group-digits=integer,group-separator={\,},group-minimum-digits=4]
    *{2}{S[table-format=6.1,table-column-width=0.112\textwidth,
      group-digits=integer,group-separator={\,},group-minimum-digits=4]}
    *{3}{S[table-format=2.2,table-column-width=0.071\textwidth]}
    *{3}{S[table-format=1.4,table-column-width=0.075\textwidth]}}
\toprule
\multirow{2}{*}{Scene}
& \multicolumn{3}{c}{Gaussian count}
& \multicolumn{3}{c}{Coverage (\%)}
& \multicolumn{3}{c}{Mean bound} \\
\cmidrule(lr){2-4}\cmidrule(lr){5-7}\cmidrule(lr){8-10}
& {Original} & {$20\%$} & {$10\%$}
& {Original} & {$20\%$} & {$10\%$}
& {Original} & {$20\%$} & {$10\%$} \\
\midrule
bicycle  & 4393644 & 878729 & 439365 & 94.95 & 97.80 & 97.94 & 0.1350 & 0.1597 & 0.1630 \\
bonsai   & 1102270 & 220454 & 110227 & 91.31 & 95.81 & 95.99 & 0.0425 & 0.0523 & 0.0545 \\
counter  & 1015704 & 203141 & 101571 & 91.67 & 95.91 & 96.08 & 0.0554 & 0.0664 & 0.0690 \\
flowers  & 2712997 & 542600 & 271300 & 92.79 & 96.38 & 96.79 & 0.1495 & 0.1755 & 0.1828 \\
garden   & 3309450 & 661890 & 330945 & 90.42 & 95.39 & 95.67 & 0.0642 & 0.0834 & 0.0880 \\
kitchen  & 1423911 & 284783 & 142392 & 91.26 & 95.76 & 96.33 & 0.0480 & 0.0594 & 0.0619 \\
room     & 1279281 & 255857 & 127929 & 90.65 & 95.38 & 95.64 & 0.0474 & 0.0566 & 0.0582 \\
stump    & 3517089 & 703418 & 351709 & 91.53 & 97.87 & 98.19 & 0.1354 & 0.1706 & 0.1750 \\
treehill & 3204008 & 640802 & 320401 & 94.68 & 97.64 & 97.84 & 0.1586 & 0.1885 & 0.1910 \\
\midrule
\rowcolor{qgsSummary}
\textbf{Average}
& 2439817.1 & 487963.8 & 243982.1
& 92.14 & 96.44 & 96.72
& 0.0929 & 0.1125 & 0.1159 \\
\bottomrule
\end{tabularx}
\end{table*}

\begin{figure*}[t]
    \centering
    \includegraphics[width=0.97\textwidth]{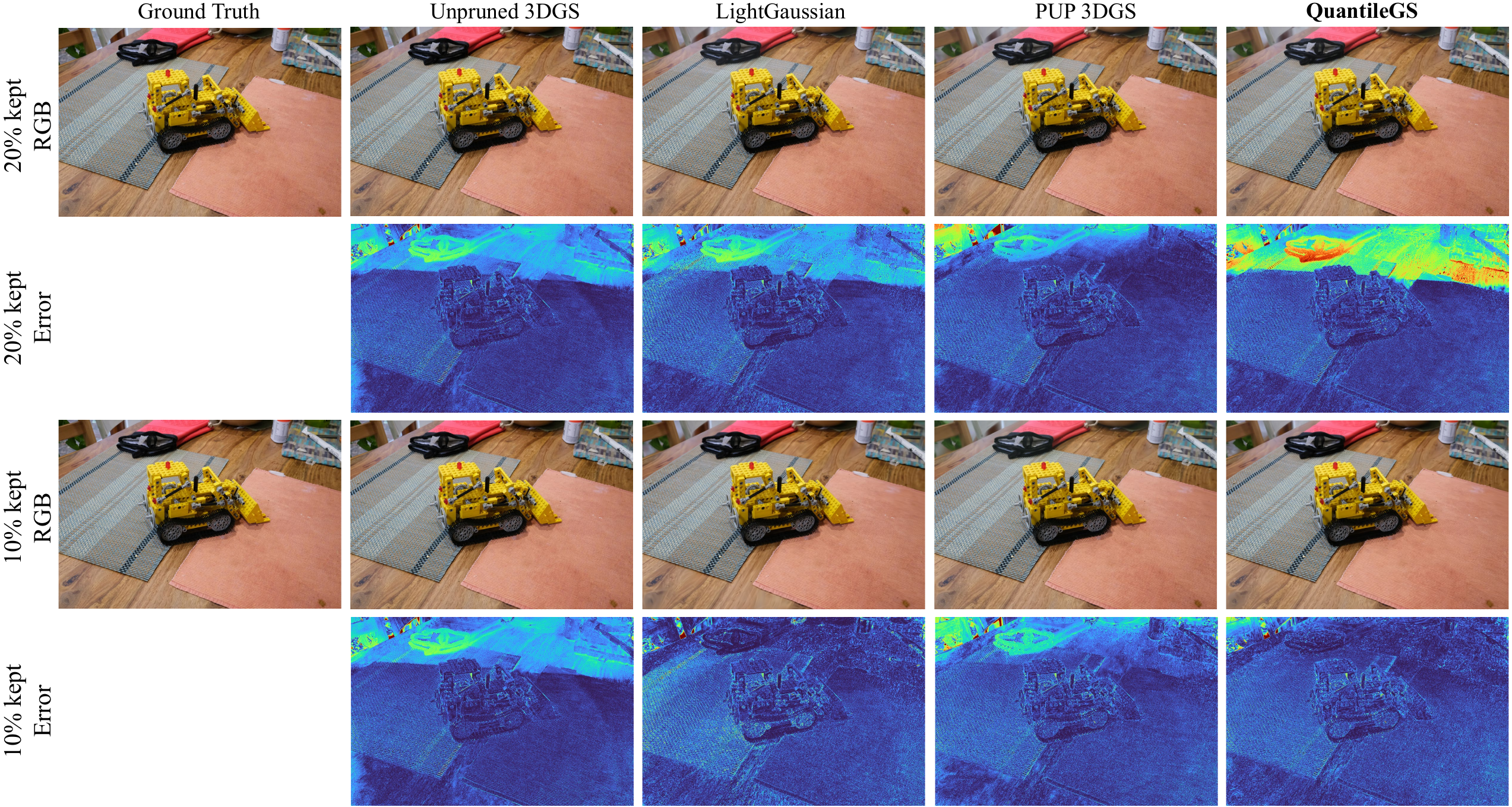}
    \par\vspace{3pt}
    \includegraphics[width=0.97\textwidth]{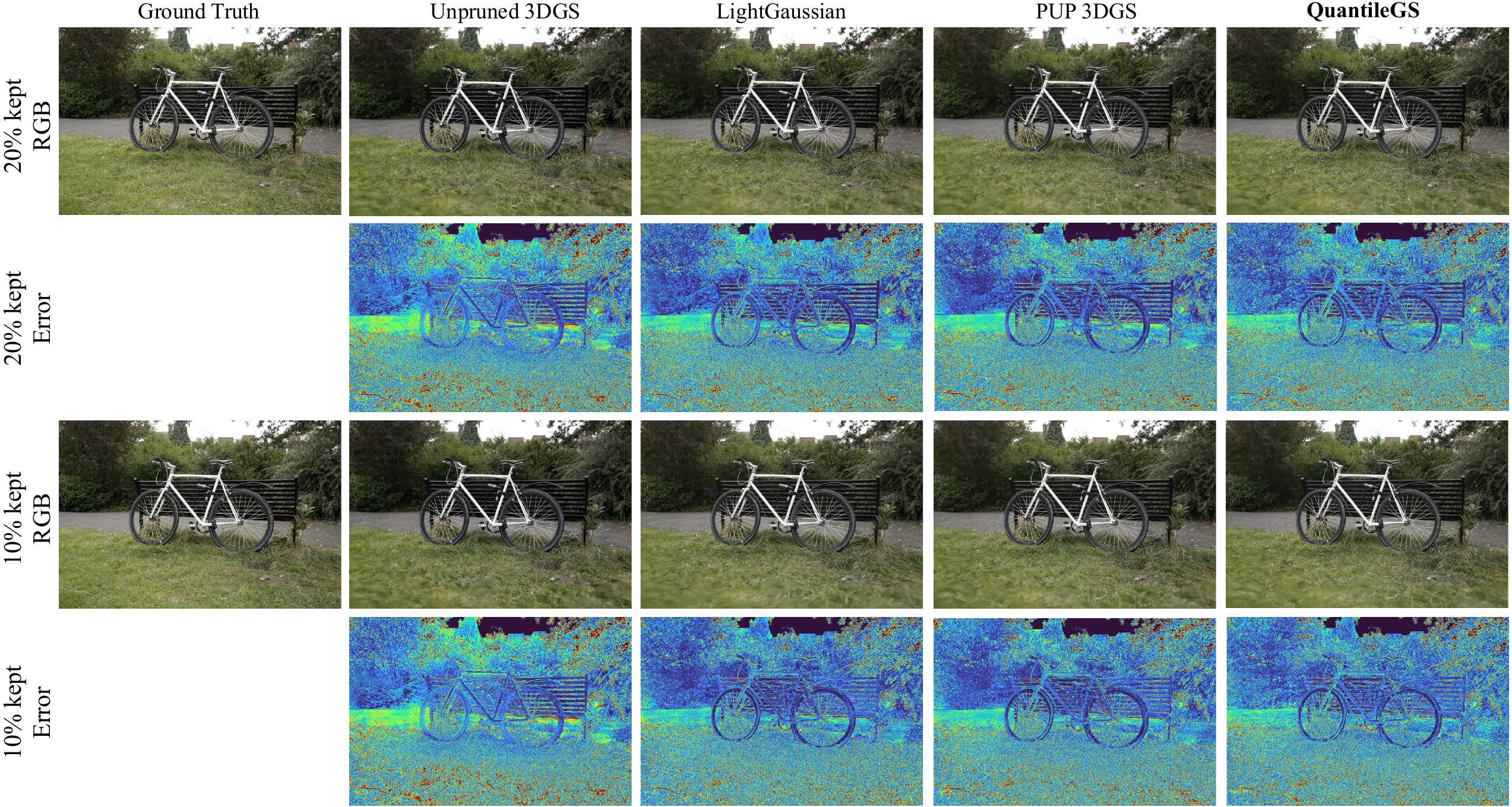}
    \caption{
        Pruning comparisons on Mip-NeRF~360 kitchen (top)
        and bicycle (bottom).
        Columns show ground truth, unpruned 3DGS,
        LightGaussian, PUP~3D-GS, and QuantileGS.
        Within each scene, the upper and lower row pairs
        show RGB renderings and RGB-MAE maps at 20\%
        and 10\% retention, respectively.
        All error maps share the $[0,0.20]$ color scale,
        with errors above 0.20 shown at the maximum color.
    }
    \label{fig:supp_pruning_comparison}
\end{figure*}

\paragraph{Qualitative comparison.}
Fig.~\ref{fig:supp_pruning_comparison} compares RGB renderings and RGB-MAE maps on the same test views of the kitchen and bicycle scenes at both retention levels. QuantileGS achieves the highest scene-level PSNR at both levels, with its largest gain over PUP occurring at 10\% retention.

\paragraph{Calibration and Target Coverage}
\label{sec:supp_calibration_targets}

We study the effect of calibration on all nine Mip-NeRF~360
scenes using the original RGB checkpoints and quantile fields from our pruning protocol, evaluated before any pruning or recovery.
The RGB renderer and the learned quantile fields remain fixed throughout this analysis.
In particular, the fitted quantile levels are $\eta=0.5$ and $\tau=0.9$ for every variant, and changing the calibration target does not involve refitting these fields. We reuse the saved fitting, calibration, and test partitions, with 21-54 calibration views per scene, and use seed~0. No test observations are used to determine the calibration corrections.

We compare the uncalibrated upper prediction, denoted by U,
with gap-normalized conformal calibration at nominal coverage levels $1-\alpha\in\{0.80,0.90,0.95\}$, denoted by C80, C90, and C95. Coverage and mean bounds are computed by pooling test pixels within each scene and then averaging scenes equally. RGB values are normalized to $[0,1]$, and bounds are expressed in RGB-MAE units. Table~\ref{tab:supp_calibration_targets} shows that the
uncalibrated prediction covers 80.01\% of test pixels on
average, below the nominal 90\% level considered by C90.
Calibration raises empirical coverage to 92.14\%, accompanied by an increase in the mean bound from 0.0513 to 0.0929. This comparison demonstrates reduced empirical undercoverage relative to the prescribed target.

Increasing the nominal target from 80\% to 90\% and 95\%
raises empirical coverage from 82.71\% to 92.14\% and 96.65\%, respectively, while the mean bound increases from 0.0593 to 0.0929 and 0.1379. These results illustrate the empirical trade-off between coverage and bound tightness:
higher coverage requirements produce more conservative bounds, without changing the fitted quantile fields.
Since calibration leaves the RGB renderer unchanged, all
variants produce identical RGB images and rendering-quality
metrics for the original model. This property does not imply unchanged performance when the bounds guide subsequent downstream decisions.

\begin{table*}[t]
\centering
\small
\captionsetup{font=small,skip=5pt,belowskip=0pt}
\setlength{\tabcolsep}{3pt}
\renewcommand{\arraystretch}{1.12}
\sisetup{detect-weight=true,detect-family=true,group-digits=none}
\begin{minipage}[t]{0.485\textwidth}
\vspace{0pt}
\caption{\textbf{Effect of the calibration target.}
Equal-scene means over nine unpruned Mip-NeRF~360 scenes
(seed~0), using the same RGB renderer, fitted quantile fields,
and test observations. C80/C90/C95 differ only in calibration
target; U has no prescribed conformal target.
The $90\%$ setting is shaded.}
\label{tab:supp_calibration_targets}

\begin{tabularx}{\linewidth}{@{}X S[table-format=2.0,table-column-width=0.24\linewidth] S[table-format=2.2,table-column-width=0.27\linewidth] S[table-format=1.4,table-column-width=0.25\linewidth]@{}}
\toprule
Variant & {Target (\%)} & {Coverage (\%)} & {Mean bound} \\
\midrule
U   & {--} & 80.01 & 0.0513 \\
C80 & 80   & 82.71 & 0.0593 \\
\rowcolor{qgsSummary}
\textbf{C90} & 90 & 92.14 & 0.0929 \\
C95 & 95   & 96.65 & 0.1379 \\
\bottomrule
\end{tabularx}
\end{minipage}%
\hfill%
\begin{minipage}[t]{0.485\textwidth}
\vspace{0pt}
\caption{\textbf{UQ and calibration in RGB-D SLAM.}
Eight-scene means for seed~0. ATE and RPE are in centimeters.
Coverage and mean RGB-MAE bounds are measured before
reference-color updates.
Darker shades indicate better tracking ranks.}
\label{tab:supp_slam_uq_ablation}

\begin{tabularx}{\linewidth}{@{}X S[table-format=1.4,table-column-width=0.17\linewidth] S[table-format=1.5,table-column-width=0.20\linewidth] S[table-format=2.2,table-column-width=0.17\linewidth] S[table-format=1.5,table-column-width=0.22\linewidth]@{}}
\toprule
\multirow{2}{*}{Method}
& \multicolumn{2}{c}{Tracking (cm)}
& \multicolumn{2}{c}{Pre-update UQ} \\
\cmidrule(lr){2-3}
\cmidrule(lr){4-5}
& {ATE $\downarrow$}
& {RPE $\downarrow$}
& {Cov. (\%)}
& {Mean bound} \\
\midrule
No-UQ
& {\third{\num{0.3427}}}
& {\third{\num{0.01021}}}
& {--}
& {--} \\
U
& {\best{\num{0.2800}}}
& {\best{\num{0.00443}}}
& 95.07
& 0.05028 \\
Online
& {\qgsSecond{\num{0.2898}}}
& {\qgsSecond{\num{0.00455}}}
& 89.59
& 0.03078 \\
\bottomrule
\end{tabularx}
\end{minipage}

\par\vspace{12pt}
\caption{\textbf{Fixed and online SLAM calibration on Replica.}
Both variants use seed~0; Online uses a step size of $0.01$.
ATE and translational RPE are RMSE values in centimeters.
Bold marks the better value in each Fixed/Online pair.
Darker shades indicate better ranks in the equal-scene
Average row.}
\label{tab:supp_slam_fixed_adaptive}

\begin{tabularx}{\linewidth}{@{}X *{2}{S[table-format=1.3,table-column-width=0.132\linewidth]} *{2}{S[table-format=1.5,table-column-width=0.132\linewidth]} *{2}{S[table-format=1.3,table-column-width=0.132\linewidth]}@{}}
\toprule
\multirow{2}{*}{Scene}
& \multicolumn{2}{c}{ATE $\downarrow$}
& \multicolumn{2}{c}{RPE $\downarrow$}
& \multicolumn{2}{c}{FPS $\uparrow$} \\
\cmidrule(lr){2-3}
\cmidrule(lr){4-5}
\cmidrule(lr){6-7}
& {Fixed} & {Online}
& {Fixed} & {Online}
& {Fixed} & {Online} \\
\midrule
room0
& 0.493 & \bfseries 0.483
& 0.00586 & \bfseries 0.00577
& 1.028 & \bfseries 1.051 \\
room1
& 0.280 & \bfseries 0.279
& 0.00512 & \bfseries 0.00505
& 1.008 & \bfseries 1.023 \\
room2
& 0.278 & \bfseries 0.275
& 0.00405 & \bfseries 0.00399
& \bfseries 1.025 & 1.014 \\
office0
& 0.372 & \bfseries 0.368
& 0.00388 & \bfseries 0.00387
& 1.103 & \bfseries 1.108 \\
office1
& 0.164 & \bfseries 0.158
& \bfseries 0.00357 & 0.00359
& \bfseries 1.096 & 1.065 \\
office2
& \bfseries 0.258 & 0.263
& \bfseries 0.00436 & 0.00441
& 1.035 & \bfseries 1.052 \\
office3
& \bfseries 0.270 & 0.272
& 0.00450 & \bfseries 0.00446
& 1.042 & \bfseries 1.051 \\
office4
& 0.227 & \bfseries 0.221
& 0.00538 & \bfseries 0.00527
& \bfseries 1.109 & 1.097 \\
\midrule
\rowcolor{qgsSummary}
\textbf{Average}
& {\qgsSecond{\num{0.293}}}
& {\best{\num{0.290}}}
& {\qgsSecond{\num{0.00459}}}
& {\best{\num{0.00455}}}
& {\qgsSecond{\num{1.056}}}
& {\best{\num{1.058}}} \\
\bottomrule
\end{tabularx}
\end{table*}

\subsection{Simultaneous Localization and Mapping}
\label{sec:supp_slam_results}

\paragraph{Role of coverage in tracking.}
Rendering-error bounds determine both reference correction budgets and photometric residual scales. For RGB-MAE, a covered error implies a per-channel error
of at most $3u_{k,p}$. Thus, at a covered fitting pixel, the budget in Eq.~\eqref{eq:slam_reference_correction} allows full correction to the observed color.
The transported bound subsequently sets the robust scale
of the photometric tracking loss. Coverage therefore provides an explicit criterion for
assessing the rendering-error bounds used by these decisions.

\begin{figure*}[!p]
    \centering
    \captionsetup{skip=4pt}

    \includegraphics[width=\textwidth]{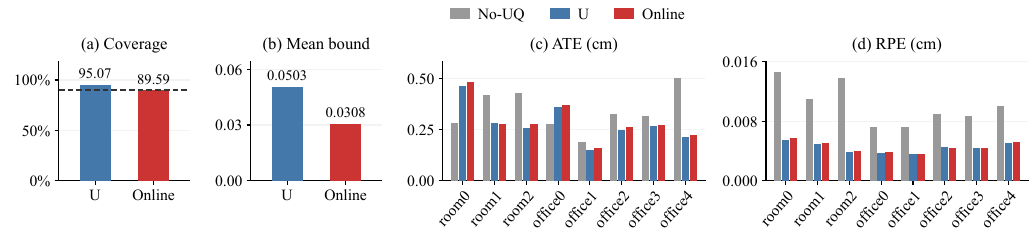}
    \caption{
        Uncertainty and calibration on eight Replica sequences.
        (a,b) Mean pre-update RGB coverage and RGB-MAE bounds.
        The dashed line marks 90\%.
        (c,d) Per-scene ATE and translational RPE in centimeters.
        Online reuses the previously evaluated runs.
    }
    \label{fig:supp_slam_uq_ablation}

    \smallskip

    \includegraphics[width=0.94\textwidth]{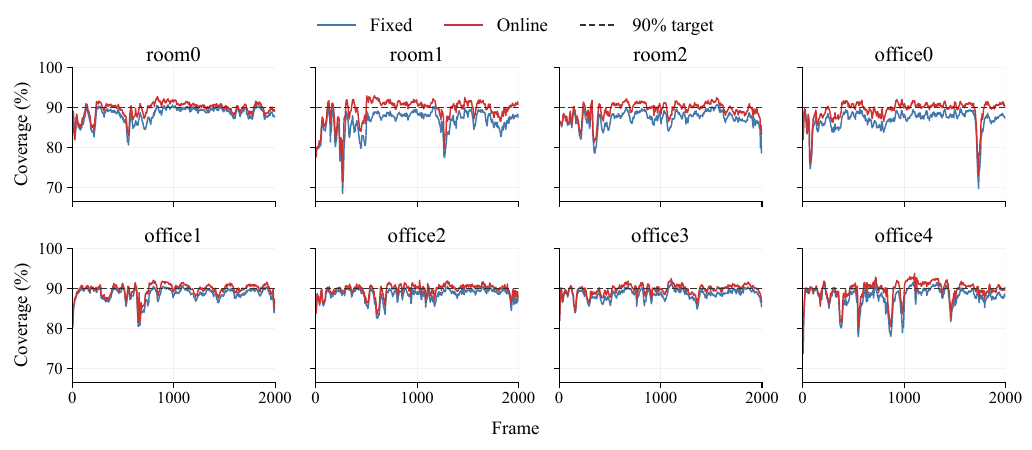}
    \caption{
        Pre-update RGB holdout coverage on eight Replica sequences.
        Blue and red denote fixed and online calibration.
        The dashed line marks 90\%.
        Each curve connects 399 evaluation frames without smoothing.
    }
    \label{fig:supp_slam_calibration}

    \smallskip

    \includegraphics[width=0.96\textwidth]{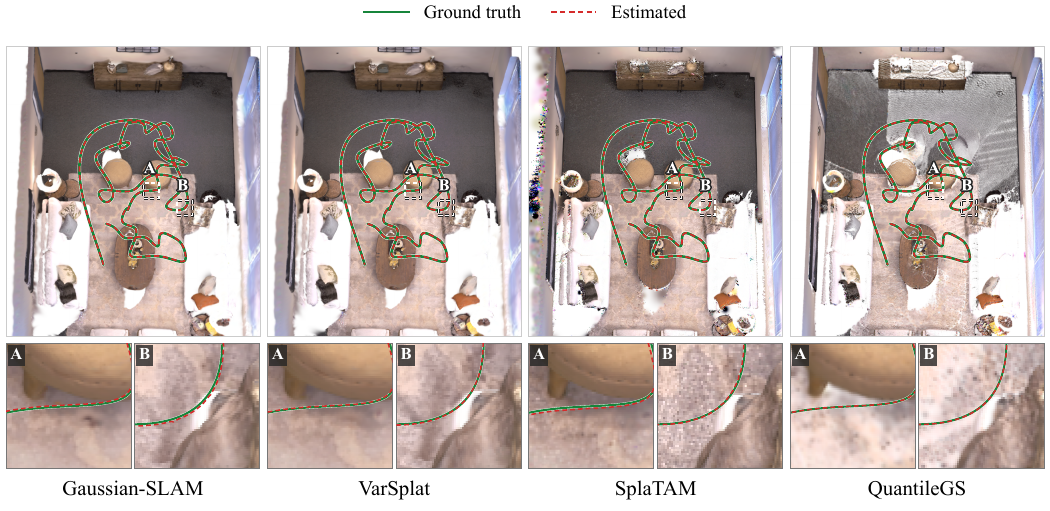}
    \caption{
        Trajectory comparison on Replica room0.
        Each column shows a method's map and shared enlarged regions A and B.
        Ground truth is green and estimated trajectories are red.
        Maps and trajectories retain their evaluated rigid alignment.
    }
    \label{fig:supp_slam_room0}
\end{figure*}

\paragraph{Uncertainty and calibration ablation.}
We compare No-UQ, which disables reference correction and uses unweighted L1 photometric tracking, with U, which uses uncalibrated bounds for reference correction and photometric residual scales. Online reuses the online-calibrated results
reported above.
Table~\ref{tab:supp_slam_uq_ablation} and
Fig.~\ref{fig:supp_slam_uq_ablation} summarize the comparison.
Relative to No-UQ, U reduces mean ATE and RPE by
18.3\% and 56.6\%, improving six and eight scenes, respectively.
This supports the joint use of uncertainty in reference
construction and tracking. Relative to U, Online reduces the mean bound by 38.8\%,
from 0.05028 to 0.03078, while coverage changes from
95.07\% to 89.59\%, closer to the nominal 90\% level
in every scene. These results show less conservative bounds with near-target
empirical coverage. However, Online has slightly higher mean ATE and RPE than U, so this improvement in bound tightness does not translate into better mean tracking accuracy.

\paragraph{Online calibration.}
We compare fixed calibration with the feedback mechanism
in Sec.~\ref{sec:slam_online_calibration}.
Fig.~\ref{fig:supp_slam_calibration} shows that feedback raises
mean coverage from 87.87\% to 89.59\%, bringing every scene
closer to 90\%.
Coverage and mean bound are measured on reserved RGB pixels
before map updates and averaged over 399 evaluated frames
per scene.
This comparison supports feedback as a mechanism for
reducing empirical undercoverage as the map evolves.
Table~\ref{tab:supp_slam_fixed_adaptive} also reports ATE
and RPE improvements on six scenes each.
Mean ATE decreases from 0.293 to 0.290\,cm, while throughput
remains approximately 1.06 FPS.

\paragraph{Trajectory visualization.}
Fig.~\ref{fig:supp_slam_room0} overlays ground-truth and estimated
trajectories on the maps, using a shared viewpoint and two
enlarged regions selected from the ground-truth trajectory.
All trajectories retain their evaluated rigid alignment,
without coordinate offsets or error amplification.
The maps provide spatial context. The evaluation concerns
camera tracking rather than global reconstruction quality.

\clearpage

